\pdfoutput=1
\PassOptionsToPackage{numbers, compress}{natbib}
\PassOptionsToPackage{backref=page}{hyperref}
\PassOptionsToPackage{table}{xcolor}
\documentclass{article}

\usepackage[preprint]{neurips_2024}

\usepackage{times}
\usepackage{latexsym}
\usepackage[utf8]{inputenc} \usepackage[T1]{fontenc}    \usepackage{hyperref}       \usepackage{url}            \usepackage{booktabs}       \usepackage{amsmath}
\usepackage{amsfonts}       \usepackage{amsthm}
\usepackage{nicefrac}       \usepackage{microtype}      \usepackage{xcolor}         \usepackage{hyperref}
\usepackage{xspace}
\usepackage{subfig}
\usepackage{tabularx}
\usepackage[most]{tcolorbox}
\usepackage{color,soul}
\usepackage{cleveref}

\newcommand{\ourmodel}[0]{G-RAG}

\newcommand{\mysubfig}[3][width=\linewidth]{
    \tcbitem\subfloat[#2]{\includegraphics[#1]{#3}}}
    
\renewcommand*{\backrefalt}[4]{\ifcase #1 No citations.\or
Cited on page #2.\else
Cited on pages #2.\fi
} 
\title{Don't Forget to Connect!\\ Improving RAG with Graph-based Reranking}

\author{Jialin Dong\\
  UCLA \\
  \And
  Bahare Fatemi\\
  Google Research
  \And
  Bryan Perozzi\\
  Google Research
  \And
  Lin F. Yang\\
  UCLA
  \And
  Anton Tsitsulin\\
  Google Research
} 
\begin{document}
\maketitle

\begin{abstract}
Retrieval Augmented Generation (RAG) has greatly improved the performance of Large Language Model (LLM) responses by grounding generation with context from existing documents.
These systems work well when documents are clearly relevant to a question context.
But what about when a document has partial information, or less obvious connections to the context? 
And how should we reason about connections between documents?
In this work, we seek to answer these two core questions about RAG generation.
We introduce \ourmodel{}, a reranker based on graph neural networks~(GNNs) between the retriever and reader in RAG.
Our method combines both connections between documents and semantic information (via Abstract Meaning Representation graphs) to provide a context-informed ranker for RAG.
\ourmodel{} outperforms state-of-the-art approaches while having smaller computational footprint.
Additionally, we assess the performance of PaLM~2 as a reranker and find it to significantly underperform \ourmodel{}.
This result emphasizes the importance of reranking for RAG even when using Large Language Models.
\end{abstract} \section{Introduction}
Retrieval Augmented Generation (RAG) \cite{siriwardhana2023improving} has brought improvements to many problems in text generation. One example is Open-Domain Question Answering~(ODQA)~\cite{voorhees-tice-2000-trec} which involves answering natural language questions without limiting the domain of the answers.
RAG merges the retrieval and answering processes, which improves the ability to effectively collect knowledge, extract useful information, and generate answers.
Even though it is successful in fetching relevant documents, RAG is not able to utilize connections between documents.
In the ODQA setting, this leads to the model disregarding documents containing answers, a.k.a.\ \emph{positive documents}, with less apparent connections to the question context.
We can identify these documents if we connect them with positive documents whose context is strongly relevant to the question context.
 
To find connections between documents and select highly relevant ones, the reranking process plays a vital role in further effectively filtering retrieved documents. A robust reranker also benefits the reading process by effectively identifying positive documents and elevating them to prominent ranking positions. When the reader's output perfectly matches one of the gold standard answers, it leads to an increase in exact-match performance metrics. Given our paper's emphasis on the reranking aspect, our performance metrics primarily focus on ranking tasks, specifically Mean Tied Reciprocal Ranking and MHits@10.
Thus, our paper focuses on using reranking to improve RAG -- as it is a fundamental bridge between the retrieval and reading processes.
\begin{figure}[tp]
    \centering
    \includegraphics[width=.75\textwidth]{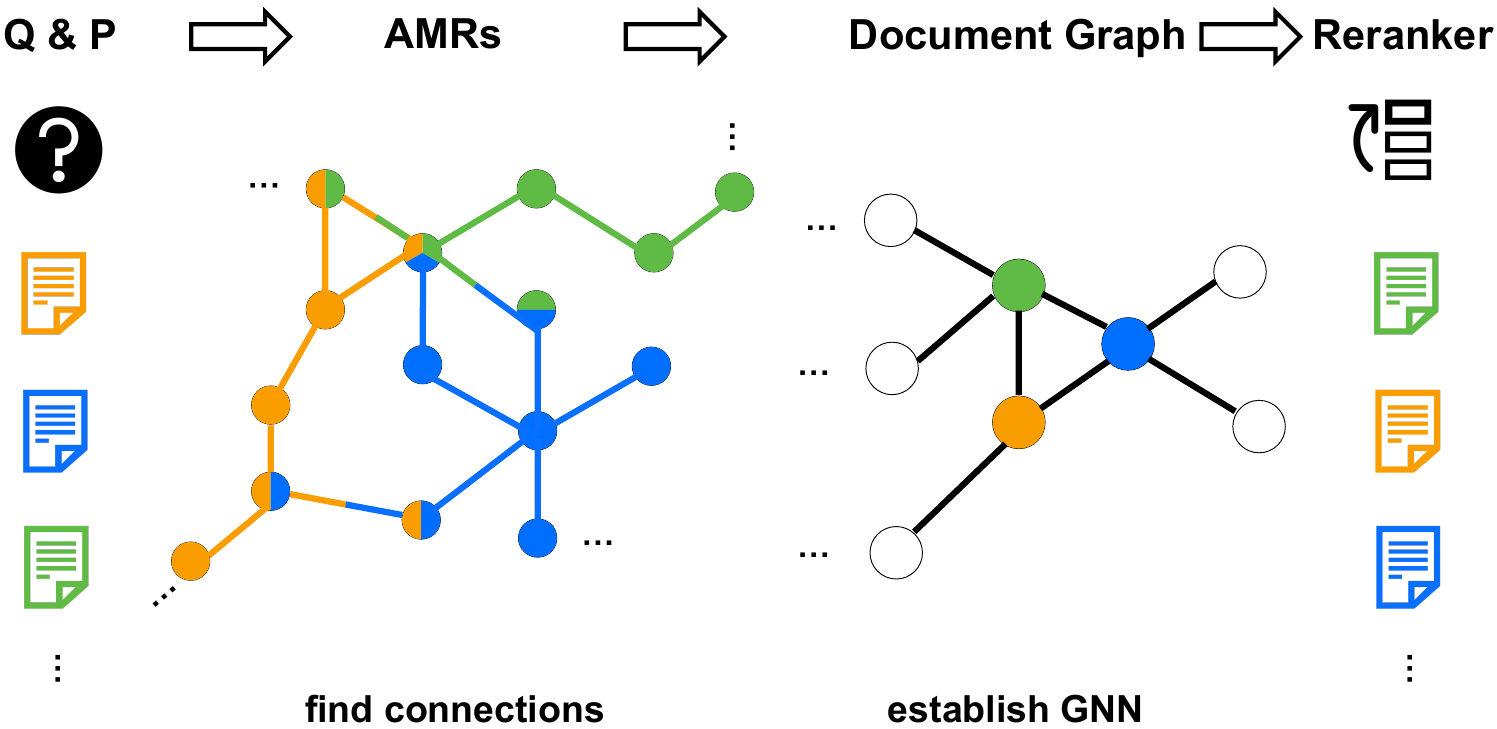}
    \caption{\ourmodel{} uses two graphs for re-ranking documents: The Abstract Meaning Representation (AMR) graph is used as features for the document-level graph. Document graph is then used for document reranking.}
\label{fig:model_framework}
    \vspace*{-4mm}
\end{figure}

Pre-trained langauge models~(LMs) like BERT \citep{BERT}, RoBERTa \citep{RoBERTa}, and BART \citep{bart} have been widely used to enhance reranking performance by estimating the relevant score between questions and documents. Recently, the Abstract Meaning Representation (AMR) graph has been integrated with a LM to enhance the system's ability to comprehend complex semantics \citep{wang-etal-2023-exploiting}. While the current rerankers exhibit admirable performance, certain limitations persist.

Firstly, as mentioned above, most of the current works fail to capture important connections between different retrieved documents. 
Some recent work \citep{KG-FiD} tries to incorporate external knowledge graphs to improve the performance of the reading process in RAG but at the cost of significant memory usage for knowledge graph storage. The connection between documents has not been considered in the reranking process yet.
Secondly, even though the AMR graph improves the understanding of the complex semantics, state-of-the-art \citep{wang-etal-2023-exploiting} work integrates redundant AMR information into the pre-trained language models.
This extra information can cause potential overfitting, in addition to increases of in computational time and GPU cost.
Thirdly, current papers utilize common pre-trained language models as rerankers which are insufficient given the fast pace of LLM development. 
With the recent breakthroughs from LLM, researchers are curious about how LLMs without perform (without fine-tuning) on the reranking task.

To address these challenges and limitations, we propose a method based on document graphs, where each node represents a document, and each edge represents that there are common concepts between two documents. We incorporate the connection information between different documents into the edge features and update the edge features through the message-passing mechanism. For node features, even though we aim to add AMR information to compose a richer understanding of complex semantics, we won't overwhelmingly add all AMR-related tokens as node-level features. Instead, we investigate the determining factor that facilitates the reranker to identify more relevant documents and encode this key factor to node features.
 
Moreover, instead of using the cross-entropy loss function during the training, we apply pairwise ranking loss in consideration of the essential aim of ranking.
We also investigate the performance of a publicly available LLM, i.e., PaLM~2~\citep{anil2023palm} with different versions, as a reranker on an ODQA. According to the moderate performance of PaLM~2 on reranking tasks, we provide several potential reasons and emphasize the irreplaceable role of reranker model design to improve RAG. The framework of graph-based reranking in the proposed \ourmodel{} is illustrated in Fig \ref{fig:model_framework}.To provide a clearer illustration of our method's pipeline, please refer to Fig \ref{fig:model_pipeline} in the Appendix.

Our contributions can be summarized as follows:
\begin{enumerate}
    \item To improve RAG for ODQA, we propose a document-graph-based reranker that leverages connections between different documents. When the documents share similar information with their neighbor nodes, it helps the reranker to successfully identify the documents containing answer context that is only weakly connected to the question.
   
\item We introduce new metrics to assess a wide range of ranking scenarios, including those with tied ranking scores. The metrics effectively evaluate this scenario by diminishing the optimistic effect brought by tied rankings. Based on these metrics, our proposed method outperforms state-of-the-art and requires fewer computational resources.
    \item We assess the performance of 
a publicly available LLM (PaLM~2~\citep{anil2023palm}) as a reranker, exploring variations across different model sizes. We find that excessive ties within the generated ranking scores hinder the 
effectiveness of pre-trained large language models in improving RAG through reranking.
\end{enumerate} \section{Related Work}\label{sec:related-work}

\paragraph{RAG in ODQA.}
RAG \cite{lewis2020retrieval,siriwardhana2023improving} combines information retrieval (via Dense Passage Retrieval, DPR \cite{DPR}) and a reading process in a differentiable manner for ODQA. 
A line of literature focuses on developing rerankers for further improving RAG. Approaches like monoT5 \citep{nogueira2020document} and monoELECTRA \citep{pradeep2022squeezing} use proposed pre-trained models. Moreover, \citet{zhuang2023rankt5} propose a fine-tuned T5 version as a reranker. More recently, \citet{park2023rink} develop a reranker module by fine-tuning the reader's neural networks through a prompting method. However, the above approaches neglect to investigate the connections among documents and fail to leverage this information during the reranking process. These methods are prone to fail to identify the documents containing gold answers that may not exhibit obvious connections to the question context. To address this issue, our proposed method is based on document graphs and is more likely to identify valuable information contained in a document if most of its neighboring document nodes in the graph share similar information.
\paragraph{Graphs in ODQA.}
Knowledge graphs, which represent entities and their relations, have been leveraged in ODQA \citep{KG-FiD,GRAPE,pathretriever,costa2018leveraging} to improve the performance of RAG. However, KG-based methods require large external knowledge bases and entity mapping from documents to the entities in the knowledge graph, which would increase the memory cost. Our proposed method does not depend on external knowledge graphs. 
While recent work by \citet{wang-etal-2023-exploiting} uses AMR graphs generated from questions and documents to construct embeddings, their focus remains on text-level relations within single document. In contrast, our approach uniquely leverages document graphs to characterize cross-document connections, a novel application within the RAG reranking process. 

\paragraph{Abstract Meaning Representation (AMR).} AMR \citep{AMR} serves as a promising tool for representing textual semantics through a rooted, directed graph. In the AMR graph, nodes represent basic semantic units like entities and concepts, while edges denote the connections between them. AMR graphs have more structured semantic information compared to the general form of natural language \citep{amr4dial, DocAMR}. A line of literature has integrated AMR graphs into learning models. Recently, \citet{wang-etal-2023-exploiting} have applied AMR to ODQA to deal with complex semantic information. Even though the performance of the reranker and the reader is improved in \citep{wang-etal-2023-exploiting}, their method also increases the computational time and GPU memory cost. This issue may arise by integrating all tokens of AMR nodes and edges without conscientiously selecting the key factors. To address this issue, our method aims to investigate the graph structure of AMR graphs and identify the key factors that improve the performance of the reranker.

\paragraph{LLMs in Reranking.}
LLMs such as ChatGPT~\citep{Chatgpt}, PaLM~2~\citep{anil2023palm}, LLaMA~\citep{touvron2023llama}, and GPT4~\citep{gpt4}, have proven to be capable of providing answers to a broad range
of questions due to their vast knowledge repositories and chain-of-thought reasoning capability. With this breakthrough, researchers are seeking to explore potential improvements that LLMs
can bring to improve RAG in ODQA, such as \citep{huang2023dsqa,ma2023fine}. At the same time, several studies \citep{wang2023evaluating, tan2023can} have scrutinized the efficacy of LLMs in Question-Answering. \citet{wang2023evaluating} indicates the superiority of the DPR \citep{DPR} + FiD \citep{FiD} approach over LLM in ODQA. While some papers have demonstrated improvements in LLM reranking performance, it's essential to note that these enhancements often involve additional techniques such as augmented query generation \cite{sun2023chatgpt} or conditional ranking tasks \cite{hou2024large}, which may not directly align with our zero-shot setting. The recent paper \cite{ma2023large} demonstrates that LLM is a good few-shot reranker and investigates different scenarios where zero-shot LLMs perform poorly. It also provides efforts to address these challenges by combining various techniques, such as employing smaller language models. Despite these investigations, the potential of LLMs without fine-tuning as rerankers to improve RAG remains unexplored, as existing studies often take pre-trained language models such as BERT \citep{BERT}, RoBERTa \citep{RoBERTa}, and BART \citep{bart} in the reranker role.

 \section{Proposed Method: \ourmodel{}}\label{sec:method}
\ourmodel{} leverages the rich structural and semantic information provided by the AMR graphs to enhance document reranking. \Cref{sec:amr} details how we use AMR graph information and build a graph structure among the retrieved documents. 
\Cref{sec:gnn} outlines the design of our graph neural network architecture for reranking documents.
\subsection{Establishing Document Graphs via AMR}\label{sec:amr}
In ODQA datasets we consider, one \textit{document} is a text block of 100 words that come from the text corpus. For each question-document pair, we concatenate the question $q$ and document $p$ as $``\text{question:<question text><document text>}"$ and then exploit AMRBART \citep{AMRBART} to parse the sequence into a singular AMR graph. The AMR graph for question $q$ and document $p$ is denoted as $G_{qp} = \{V, E\}$, where $V$ and $E$ are nodes and edges, respectively. Each node is a concept, and each edge is denoted as $e = (s, r, d)$ where $s, r, d$ represent the source node, relation, and the destination node, respectively. Our reranker aims to rank among the top 100 documents retrieved by DPR~\citep{DPR}. Thus, given one question $q$ and documents $\{p_1,\cdots,p_{n}\}$ with $n=100$, we establish the undirected document graph $\mathcal{G}_q=\{\mathcal{V},\mathcal{E}\}$ based on AMRs $\{G_{qp_1}, \cdots,G_{qp_{n}}\}$. For each node $v_i\in\mathcal{V}$, it corresponds to the document $p_i$. 
For $v_i, v_j\in\mathcal{V}$, $i\neq j$, if the corresponding AMR $G_{qp_i}$ and $G_{qp_j}$ have common nodes, there will be an undirected edge between $v_i$ and $v_j$ (with a slight abuse in notation) denoted as $e_{ij} = (v_i,v_j)\in \mathcal{E}$. We remove isolated nodes in $\mathcal{G}_q$. In the following, 
we will construct the graph neural networks based on the document graphs to predict whether the document is relevant to the question. Please refer to Appendix \ref{appexdix:data} for AMR graph statistics, i.e., the number of nodes and edges in AMR graphs, of the common datasets in ODQA.

\subsection{Graph Neural Networks for Reranking}\label{sec:gnn}
Following Section \ref{sec:amr}, we construct a graph among the $n=100$ retrieved documents denoted as $\mathcal{G}_q$ given the question $q$. We aim to exploit both the structural information and the AMR semantic information to rerank the retrieved documents. To integrate the semantic information of documents, the pre-trained language models such as BERT \citep{BERT}, and RoBERTa \citep{RoBERTa} are powerful tools to encode the document texts as node features in graph neural networks. Even though \citet{wang-etal-2023-exploiting} integrate AMR information into LMs, it increases computational time and GPU memory usage. To address this, we proposed node and edge features for graph neural networks, which simultaneously exploit the structural and the semantic information of AMR but avoid adding redundant information.

\subsubsection{Generating Node Features} Our framework applies a pre-trained language model to encode all the $n$ retrieved documents in $\{p_1, p_2,\cdots, p_n\}$ given a question $q$. The document embedding is denoted as $ \tilde{X} \in \mathbb{R}^{n\times d}$ where $d$ is the hidden dimension, and each row of $ \tilde{X}$ is given by
\begin{align}
 \tilde{x}_i & = \mathrm{Encode}(p_i) \text{  for  } i \in \{1,2,\cdots n\}.\label{eq:feat_base}
\end{align}

Since AMR brings more complex and useful semantic information, we intend to concatenate document text and corresponding AMR information as the input of the encoder. However, if we integrate all the information into the embedding process as the previous work \citep{wang-etal-2023-exploiting} did, it would bring high computational costs and may lead to overfitting. To avoid this, we investigate the determining factor that facilitates the reranker to identify more relevant documents. By studying the structure of AMRs for different documents, we note that almost every AMR has the node ``question'', where the word ``question" is included in the input of the AMR parsing model, given by $``\text{question:<question text><document text>}"$. Thus, we can find the single source shortest path starting from the node ``question". When listing every path, the potential connection from the question to the answer becomes much clearer. By looking into the nodes covered in each path, both the structural and semantic information can be collected. The embedding enables us to utilize that information to identify the similarity between question and document context. 

To better illustrate the structure of the shortest path, we also conduct some experiments to show the statistic of the shortest path, see Fig \ref{fig:shortest path} in Appendix. We study the shortest single source paths (SSSPs) starting from ``question'' in the AMR graphs of documents from the train set of Natural Question (NQ) \citep{kwiatkowski2019natural} and TriviaQA (TQA)\citep{joshi-etal-2017-triviaqa} dataset. The analysis shows
that certain negative documents cannot establish adequate connections to the question context within their text. Moreover, negative documents encounter another extreme scenario where paths contain an abundance of information related to the question text but lack valuable information such as the gold answers. This unique pattern provides valuable insight that can be utilized during the encoding process to improve the reranker performance.

Thus, the proposed document embedding is given by $X \in \mathbb{R}^{n\times d}$ and each row of $X$ can be given by, for $i \in \{1,2,\cdots n\}$:
\begin{align}
 x_i & = \mathrm{Encode}(\mathrm{concat}(p_i,a_i)),\label{equ:input_node_feat}
\end{align}
where $a_i$ is a sequence of words, representing the AMR information concerning the document $p_i$. There are two steps to get the representation of $a_i$: 1) Path Identification: Firstly, the shortest single source paths (SSSPs) are determined starting from the node labeled "question" in the AMR graph $G_{qp_i}$. Each path identified should not be a subset of another. For instance, consider the following paths composed of node concepts: [`question', `cross', `world-region', `crucifix', `number', `be-located-at', `country', `Spain'], [`question', `cross', `religion', `Catholicism', `belief', `worship'];
2) Node Concept Extraction: Subsequently, the node concepts along these identified paths are extracted to construct $a_i$. In the example provided, $a_i$ is formed as follows: "question cross world-region crucifix number be-located-at country Spain religion Catholicism belief worship". $X \in \mathbb{R}^{n\times d}$ (\ref{equ:input_node_feat}) will be the initial node representation of graph neural networks.

\subsubsection{Edge Features} Besides the node features, we also adequately leverage edge features associated with undirected edges in AMR $\{G_{qp_1}, \cdots, G_{qp_{n}}\}$. Let $\hat{{E}}\in\mathbb{R}^{n\times{}n\times{}l}$ denote the edge features of the graph. Then, $\hat{{E}}_{ij\cdot}\in{}\mathbb{R}^l$ represents the $l$-dimensional feature vector of the edge between the node $v_i$ and node $v_j$ $i \neq j$, and $\hat{{E}}_{ijk}$ denotes the $k$-{th} dimension of the edge feature in $\hat{{E}}_{ij\cdot}$. In our framework, $l=2$ and $\hat{{E}}$ is given by:
\begin{align}
\left\{
\begin{array}{ll}
      \hat{{E}}_{ij\cdot}={0},  \text{no connection between $G_{qp_i}$ and $G_{qp_j}$}, \\
      \hat{{E}}_{ij1}= \text{\# common nodes between $G_{qp_i}$ and $G_{qp_j}$}, \\
      \hat{{E}}_{ij2}= \text{\# common edges between $G_{qp_i}$ and $G_{qp_j}$}.\\
\end{array}
\right. 
\end{align}
We then normalize the edge feature $\hat{{E}}$ to avoid
the explosive scale of output node features when being multiplied by the edge feature in graph convolution operations. Thus, our derived feature ${E}$ is normalized on the first and second dimension, respectively.
Similar edge normalization has also been considered in the paper \citep{gong2019exploiting}. $E\in\mathbb{R}^{n\times{}n\times{}l}$ will be the initial edge representation of graph neural networks.

\subsubsection{Representation Update} Based on the above initial node and edge representations, we arrive at updating representations in the graph neural networks. Given a document graph $\mathcal{G(\mathcal{V},\mathcal{E})}$ with $|\mathcal{V}|=n$,
the input feature of node $v\in\mathcal{V}$ is denoted as $\mathbf{x}_v^{0}\in\mathbb{R}^d$, and the initial representation of the edge
between node $v$ and $u$ is given by $\mathbf{e}_{uv}^0\in\mathbb{R}^l$ with $l=2$. Let $\mathcal{N}(v)$ denote the neighbor nodes of the node $v\in\mathcal{V}$.
The representation of node $v\in\mathcal{V}$ at layer $\ell$ can be derived from a GNN model given by:
\begin{equation}\label{eq:mp-vertex}
\mathbf{x}_v^{\ell} = g(\mathbf{x}_v^{\ell-1},\bigcup_{u\in\mathcal{N}(v)} f(\mathbf{x}_u^{\ell-1},  \mathbf{e}_{uv}^{\ell-1})),
\end{equation}
where $f$, $\bigcup$ and $g$ are functions for computing feature, aggregating data, and
updating node representations, respectively. Specifically, the function $f$ applies different dimensional edge features as weights to the node features, given by
\begin{align}
    f(\mathbf{x}_u^{\ell-1},  \mathbf{e}_{uv}^{\ell-1}) = \sum_{m=1}^{l}{e}_{uv}^{\ell-1}(m)\mathbf{x}_u^{\ell-1}.
\end{align}
We choose mean aggregator \citep{kipf2017semi} as the operation $\bigcup$. The parameterized function $g$ is a non-linear learnable function that aggregates the representation of the node and its neighbor nodes.
Simultaneously, the representation of edge starting from $v\in\mathcal{V}$ at layer $\ell$ is given by:
\begin{equation}\label{eq:mp-edge}
\mathbf{e}_{v\cdot}^{\ell} = g(\mathbf{e}_{v\cdot}^{\ell-1},\bigcup_{u\in\mathcal{N}(v)} \mathbf{e}_{u\cdot}^{\ell-1}).
\end{equation} 

\subsubsection{Reranking Score and Training Loss} Given a question $q$ and its document graph $\mathcal{G}_q=\{\mathcal{V},\mathcal{E}\}$, we have the output node representations of GNN, i.e., $\mathbf{x}_v^{L}$, where $L$ is the number of GNN layers. With the same encoder in (\ref{equ:input_node_feat}), the question $q$ is embedded as 
\begin{align}
    \mathbf{y} = \mathrm{Encode}(q).
\end{align}
The reranking score for each node $v_i\in\mathcal{V}$ corresponding the document $p_i$ is calculated by 
\begin{align}
s_i = \mathbf{y}^{\top}\mathbf{x}_{v_i}^{L},\label{eq:relevant_score}
\end{align}
 for $i = 1,\cdots,n$ and $|\mathcal{V}|=n$. The cross-entropy training loss of document ranking for the given question $q$ is:
\begin{align}
    \mathcal{L}_{q} = - \sum_{i=1}^{n} y_i \log \left( \frac{\exp(s_i)}{\sum_{j=1}^{n}\exp(s_j)} \right)
\label{eq:loss}
\end{align}
where $y_i=1$ if $p_{i}$ is the positive document, and $0$ for the negative document. The cross-entropy loss may fail to deal with the unbalanced data in ODQA where the number of negative documents is much greater than the number of positive documents.
Besides the cross-entropy loss function, the pairwise loss function has been a powerful tool for ranking \cite{li2017improving}. Given a pair of scores $s_i$ and $s_j$, the ranking loss is given by :
\begin{align}
    \mathcal{RL}_{q}(s_i, s_j, r) = \max\left(0, -r\left(s_i-s_j\right) + 1\right),\label{eq:ranking__loss}
\end{align}
where $r=1$ if document $i$ should be ranked higher than document $j$, and vice-versa for $r=-1$. We conduct experiments based on both loss functions and emphasize the advantage of the ranking loss (\ref{eq:ranking__loss}) over the cross-entropy loss (\ref{eq:loss}).

 \section{Experiments}\label{sec:experiments}
\subsection{Setting}\label{sec:setting}
\paragraph{Datasets.}
We conduct experiments on two representative ODQA datasets Natural Questions~(NQ)~\citep{kwiatkowski2019natural} and and TriviaQA~(TQA)~\citep{joshi-etal-2017-triviaqa}. NQ is derived from Google Search Queries and TQA includes questions from trivia and quiz-league websites. Detailed dataset statistics are presented in Table~\ref{table:dataset} in Appendix \ref{appexdix:data}. Note that the gold answer lists in dataset NQ usually have much fewer elements than the dataset TQA, which leads to a much smaller number of positive documents for each question.

We use DPR~\citep{DPR} to retrieve $100$ documents for each question and generate the AMR graph for each question-document pair using  AMRBART \citep{AMRBART}. The dataset with AMR graphs is provided by \citep{wang-etal-2023-exploiting}\footnote{\url{https://github.com/wangcunxiang/Graph-aS-Tokens/tree/main}}. 
Please refer to Appendix \ref{appexdix:data} for more details on the AMR statistic information. 
We conducted our experiments on a Tesla A100 40GB GPU, demonstrating the low computational needs of \ourmodel{}.

\paragraph{Model Details.}
For the GNN-based reranking models, we adopt a 2-layer Graph Convolutional Network \citep{kipf2017semi} with hidden dimension chosen from $\{8, 64, 128\}$ via hyperparameter-tuning. The dropout rate is chosen from $\{0.1, 0.2,0.4\}$. 
We initialize the GNN node features using pre-trained models, e.g, BERT \cite{BERT}, GTE \cite{li2023general}, BGE \cite{bge_embedding}, Ember \cite{liu2022retromae}. 
We base our implementaion of the embedding model on the HuggingFace Transformers library~\citep{wolf2019huggingface}. For training our framework, we adopt the optimizer AdamW~\citep{adamw} with the learning rate chosen from $\{5e-5,1e-4,5e-4\}$. Batch size is set to $5$. We set the learning rate warm-up with $1,000$ steps. The number of total training steps is $50$k, and the model is evaluated every $10$k steps.

\subsubsection{Metrics} Even though Top-K accuracy, where the ground-truth ranking is based on DPR scores \citep{DPR}, is commonly used in the measurement of reranking \citep{Re2G,FiD}, this metric is unsuitable for indicating the overall reranking performance for all positive documents. Moreover, with the promising development of LLM in learning the relevance between texts, DPR scores may lose their advantage and fairness. To address this issue, other metrics such as Mean Reciprocal Rank (MRR) and Mean Hits@10 (MHits@10) are used for measuring the reranking performance \cite{wang-etal-2023-exploiting}. To be specific,
The Mean Reciprocal Rank (MRR) score of positive document is given by
$      \mathrm{MRR} = \frac{1}{|\mathcal{Q}|} \sum_{q \in \mathcal{Q}} (\frac{1}{|\mathcal{P}^{+}|}{\sum_{p\in \mathcal{P}^{+}}\frac{1}{r_p}}),
$
where $\mathcal{Q}$ is the question set from the evaluating dataset, $\mathcal{P}^{+}$ is the set of positive documents, and $r_p$ is the rank of document $p$ estimated by the reranker.
The MHits@10 indicates the percentage of positive documents that are ranked in the Top 10, given by
$
     \mathrm{MHits@10} = \frac{1}{|\mathcal{Q}|} \sum_{q \in \mathcal{Q}} (\frac{1}{|\mathcal{P}^{+}|}\sum_{p\in \mathcal{P}^{+}} \mathbb{I}(r_p<=10)), 
$
where the indication $\mathbb{I}(A)=0$ if the event $A$ is true, otherwise 0.

The above metrics work well for most cases, however, they may fail to fairly characterize the ranking performance when there are ties in ranking scores, which is common in relevant scores generated by LLMs such as  ChatGPT \citep{Chatgpt}, PaLM~2 \citep{anil2023palm},  LLaMA \citep{touvron2023llama}, and GPT4 \citep{gpt4}. Please refer to Fig \ref{fig:ulm_example} in the Appendix for the detailed prompt and results of relevant scores between questions and documents. To address ties in the ranking scores, we propose variants of MRR and MHits@10. Denote $r_p^{(t)}$ as the rank of the document $p$ with $t$ ties. In other words, the relevant score between the question and the document $p$ is the same as other $t-1$ documents. The variant of MRR for tied ranking is named Mean Tied Reciprocal Ranking (MTRR), represented as
\begin{align}\label{mtrr}
  \mathrm{MTRR}& = \frac{1}{|\mathcal{Q}|} \sum_{q \in \mathcal{Q}} \bigg(\frac{1}{|\mathcal{P}^{+}|}{\sum_{p\in \mathcal{P}^{+}}\frac{1}{r_p^{(t)}}}\mathbb{I}(t=1)\notag\\
  &+\frac{2}{r_p^{(t)} + r_p^{(t)} +t-1}\mathbb{I}(t>1)\bigg).
\end{align}
The metric MTRR addresses the tied rank $r_p^{(t)}$ estimated by the reranker via averaging the optimistic rank $r_p^{(t)}$ and the pessimistic rank $r_p^{(t)} +t-1$. The metrics MRR and MTRR are the same when there is no ranking tie. The variant of MHits@10 for tied ranking is Tied Mean Hits@10 (TMHit@10). Denote $\mathcal{H}(p)$ as the set that includes all the ranks $\{r_{p_i}^{(t_i)}\}$ that are higher than the rank of document $p$, i.e., $r_{p}^{(\tau)}$. Based on these notations, we present the new metric as:
\begin{align}\label{tmhit}
\mathrm{TMHits@10} =\frac{1}{|\mathcal{Q}|} \sum_{q \in \mathcal{Q}} \left(\frac{1}{|\mathcal{P}^{+}|}\sum_{p\in \mathcal{P}^{+}} \mathrm{Hits}@10(p)\right),
\end{align}
where $\mathrm{Hits}@10(p) $ is defined as
\begin{align}
\left\{
\begin{array}{ll}
 0,                       \text{if $\sum_{i} t_i>10$ for $\forall ~r_{p_i}^{(t_i)}\in \mathcal{H}(p)$},\\
     ({10-\sum_{i} t_i})/{\tau}, \text{if $0<\sum_{i} t_i<10$ }\\ \quad\text{for $\forall ~r_{p_i}^{(t_i)}\in \mathcal{H}(p)$ and $\tau>1$}, \\
      {10}/{\tau},  \text{if $\mathcal{H}(p) = \emptyset$ and $\tau>10$}, \\
     1,                       \text{otherwise}.
\end{array}
\right. \notag
\end{align}
If there are ties in the Top-10 ranking, the metric TMHit@10 diminishing the optimistic effect via dividing the hit-number (no greater than 10)  by the number of ties.
\subsection{Comparing Reranker Systems}\label{sec:exp_baseline}
We compare our proposed algorithm with baselines as follows with fixed hyper-parameters and no fine-tuning, where the hidden dimension is 8, the dropout rate is 0.1, and the learning rate is 1e-4. \textbf{w/o reranker:} Without an additional reranker, the ranking score is based on the retrieval scores provided by DPR \citep{DPR}.
\textbf{BART:} The pre-trained language model BART \citep{bart} serves as the reranker. 
\textbf{BART-GST:} This method integrates graph-as-token into the pre-trained model \citep{wang-etal-2023-exploiting}. For each dataset, we use the best performance provided in the paper.
\textbf{RGCN-S}\citep{wang-etal-2023-exploiting} stacks the RGCN model on the top of the transformer. Even though this method is based on graph neural networks, it doesn't rely on the document graphs, but construct nodes in the graph model based on the text alignment in question-document pairs. \textbf{MLP:} The initial node features are only based on document text as described in (\ref{eq:feat_base}) with the BERT \citep{BERT} encoder. After the node features go through MLP, we get the relevant scores via (\ref{eq:relevant_score}) and take the cross-entropy function (\ref{eq:loss}) as training loss.
 \textbf{GCN:} Besides updating node representations via GCN, the rest setting is the same as MLP. We also conduct experiments with different GNN models. Please refer to Appendix \ref{appendix:gnn} for details.
 \textbf{\ourmodel{}:} The initial node features are based on document text and AMR information as described in (\ref{equ:input_node_feat}). The rest of the setting is the same as {GCN}.
 \textbf{\ourmodel{}-RL:} Using the ranking loss function \label{eq:ranking_loss} and keep the other setting the same as  {\ourmodel{}}.

\begin{table}[ht]
\centering
\newcolumntype{C}{>{\centering\arraybackslash}X}
\begin{tabularx}{\textwidth}{@{}p{2cm}CCCCCCCC@{}}
\toprule
& \multicolumn{4}{c}{\textbf{NQ}} & \multicolumn{4}{c}{\textbf{TQA}} \\
\cmidrule(r){2-5}\cmidrule(l){6-9}
strategy &  \multicolumn{2}{c}{\textbf{MRR}} & \multicolumn{2}{c}{\textbf{MH}} & \multicolumn{2}{c}{\textbf{MRR}} & \multicolumn{2}{c}{\textbf{MH}} \\ 
\cmidrule(lr){2-3}\cmidrule(lr){4-5}\cmidrule(lr){6-7}\cmidrule(lr){8-9}
\mbox{w/o reranker} & 20.2&18.0 & 37.9&34.6 & 12.1&12.3 & 25.5&25.9 \\ 
{ BART} & 25.7&23.3 & 49.3&45.8 & 16.9&17.0 & 37.7&38.0 \\ 
{ BART-GST} & \textbf{28.4}&25.0 & \textbf{53.2}&\textbf{48.7} & 17.5&17.6 & 39.1&\textbf{39.5} \\ 
\mbox{RGCN-S} & 26.1&23.1 & 49.5&46.0 & {---} & --- & --- & --- \\ 
MLP & 19.2&17.8 & 40.0&38.8 & 17.6&17.1 & 34.0&31.4 \\ 
GCN & 22.6&22.4 & 47.6&44.2 & 18.2&17.4 & 38.0&37.0 \\ 
\midrule
\textbf{\ourmodel{}} & 25.1&24.2 & 49.1&47.2 & 18.5&\textbf{18.3} & 38.5&39.1 \\ 
\mbox{\textbf{\ourmodel{}-RL}} & 27.3&\textbf{25.7} & 49.2&47.4 & \textbf{19.8}&\textbf{18.3} & \textbf{42.9}&{39.4} \\
\bottomrule
\end{tabularx}
\caption{Results on the dev/test set of NQ and TQA without hyperparameter fine-tuning.}\label{table:result_baseline}
\vspace*{-4mm}
\end{table}
The results on MRR and MHits@10 on the NQ and TQA datasets are provided in Table \ref{table:result_baseline}. Note that the results on NQ always outperform the results on TQA, this is due to a smaller number of positive documents making it easy to put most of the positive documents into the Top 10. Generally speaking, TQA is a more complex and robust dataset than NQ. Models with graph-based approaches, such as GCN and \ourmodel{}, show competitive performance across metrics. These methods have advantages over the baseline models, i.e., without reranker and MLP. In conclusion, based on the simulation results, the proposed method \ourmodel{}-RL emerges as a strong model, indicating the effectiveness of graph-based strategies and the benefit of pairwise ranking loss on identifying positive documents. 
To highlight the advantages of the proposed \ourmodel{} over state-of-the-art benchmarks, we conducted experiments across various embedding models with fine-tuning parameter in the next section.

\begin{table*}[htp]
\centering
\newcolumntype{C}{>{\centering\arraybackslash}X}
\begin{tabularx}{\textwidth}{@{}p{2cm}CCCCCCCC@{}}
\toprule
& \multicolumn{4}{c}{\textbf{NQ}} & \multicolumn{4}{c}{\textbf{TQA}} \\
\cmidrule(r){2-5}\cmidrule(l){6-9}
strategy &  \multicolumn{2}{c}{\textbf{MRR}} & \multicolumn{2}{c}{\textbf{MH}} & \multicolumn{2}{c}{\textbf{MRR}} & \multicolumn{2}{c}{\textbf{MH}} \\ 
\cmidrule(lr){2-3}\cmidrule(lr){4-5}\cmidrule(lr){6-7}\cmidrule(lr){8-9}
\mbox{w/o reranker} & 20.2 & 18.0 & 37.9 & 34.6 & 12.1 & 12.3 & 25.5 & 25.9 \\
BART & {25.7} & {23.3} & \textbf{49.3} & {45.8} & {16.9} & {17.0} & {37.7} & {38.0} \\
\midrule
\mbox{PaLM~2 XS} & 14.9 & 14.0 & 34.1 & 34.2 & 11.6 & 12.5 & 29.1 & 31.6 \\
PaLM~2 L & 18.6 & 17.9 & 40.7 & 39.7 & 12.7 & 12.9 & 34.7 & 35.6 \\
\midrule
\mbox{\textbf{\ourmodel{}-RL}} & \textbf{27.3} & \textbf{25.7} & 49.2 & \textbf{47.4} & \textbf{19.8} & \textbf{18.3} & \textbf{42.9} & \textbf{39.4} \\
\bottomrule
\end{tabularx}
\caption{Results of PaLM~2 being the reranker. Small embedding models outperform LLMs in this setting. In comparison, \ourmodel{}-RL considerably improves the results compared to both language model types by leveraging connection information across documents. We use Tied Mean Hits@10.}
\label{table:ulm}
\vspace*{-4mm}
\end{table*}

\subsection{Using different LLMs as Embedding Models}
The feature encoder always plays a vital role in NLP tasks. Better embedding models are more likely to fetch similarities across contexts and help identify highly relevant context. Besides the BERT model used in the state-of-the-art reranker, many promising embedding models have been proposed recently.  To evaluate the effectiveness of different embedding models, i.e., BERT \cite{BERT}, GTE \cite{li2023general}, BGE \cite{bge_embedding}, Ember \cite{liu2022retromae}, we conduct the experiments under the same setting as \ourmodel{}-RL. The results are presented in Table \ref{table:embedding}. For convenience, we directly add two results from Section \ref{sec:exp_baseline}: BART-GST and BERT. Ember performs consistently well across all evaluations. In conclusion, Ember appears to be the top-performing model, followed closely by GTE and BGE, while BART-GST and BERT show slightly lower performance across the evaluated metrics. Thus our fine-tuning result is based on {\ourmodel{}-RL} with Ember as the embedding model. The grid search setting for hyperparameter is introduced in Section \ref{sec:setting}. We only run 10k iterations for each setting and pick up the one with the best MRR. The result with hyperparameter tuning, i.e., Ember (HPs-T), is added in Table \ref{table:embedding}. Even though BART-GST demonstrates competitive performance in some scenarios, it is prone to overfitting especially in terms of MRR on the NQ dataset. However, the proposed methods, i.e., Ember and Ember (HPs-T), are more likely to avoid overfitting and achieve the highest score across all test sets.
\begin{table}[htp]
\centering
\newcolumntype{C}{>{\centering\arraybackslash}X}
\begin{tabularx}{\textwidth}{@{}p{2cm}CCCCCCCC@{}}
\toprule
& \multicolumn{4}{c}{\textbf{NQ}} & \multicolumn{4}{c}{\textbf{TQA}} \\
\cmidrule(r){2-5}\cmidrule(l){6-9}
embedding &  \multicolumn{2}{c}{\textbf{MRR}} & \multicolumn{2}{c}{\textbf{MH}} & \multicolumn{2}{c}{\textbf{MRR}} & \multicolumn{2}{c}{\textbf{MH}} \\ 
\cmidrule(lr){2-3}\cmidrule(lr){4-5}\cmidrule(lr){6-7}\cmidrule(lr){8-9}
\mbox{BART-GST} & 28.4&25.0 & \textbf{53.2} & {48.7} & 17.5 & 17.6 & 39.1 & 39.5 \\
BERT & 27.3 & 25.7 & 49.2 & 47.4 & 19.8 & 18.3 & 42.9 & 39.4 \\
GTE & \textbf{29.9}&26.3 & 52.6&47.7 & 19.2&19.3 & 41.8&40.3 \\ 
BGE & 28.7&{27.4} & 52.1 & 48.2 & 18.7&18.3 & 43.4& 40.7 \\ 
\midrule
\textbf{Ember} & 9.0& 26.1 & 52.9& 48.0 & {19.8}& {18.6} & \textbf{44.3} & \textbf{42.0} \\ 
\mbox{\textbf{Ember (HPs-T)}} & 28.9& \textbf{27.7} & 51.1 & \textbf{50.0} & \textbf{20.0} & \textbf{19.4} & {41.6} &{41.4} \\
\bottomrule
\end{tabularx}
\caption{\ourmodel{} with changing the embedding model.}\label{table:embedding}
\vspace*{-6mm}
\end{table}
\subsection{Investigating PaLM~2 Scores}
To evaluate the performance of large language models on the reranking task, we conduct zero-short experiments on the dev \& test sets of the NQ and TQA datasets. An example of LLM-generated relevance score is illustrated in Figure~\ref{fig:ulm_example} in the Appendix.

In general, we observe that scores generated by PaLM 2 are integers between 0 and 100 that are divisible by 5. This often leads to ties in documents rankings. To address the ties in the ranking score, we use the proposed metrics MTRR (Eq.~\ref{mtrr}) and TMHits@10 (Eq.~\ref{tmhit}) to evaluate the performance of reranker based on PaLM~2 \cite{anil2023palm}. For the convenience of comparison, we copy \texttt{w/o rerank}, {BART}, and \ourmodel{} results from Section~\ref{sec:exp_baseline}. Since there is no tied ranking provided by \texttt{w/o rerank} and {BART}, the MRR and MHits@10 have the same values as MTRR and TMhits@10, respectively.

The performance results are provided in Table \ref{table:ulm}. The results demonstrate that LLMs with zero-shot learning does not do well in reranking tasks. This may be caused by too many ties in the relevance scores, especially for small-size LLM where there are more of them. This result emphasizes the importance of reranking model design in RAG even in the LLM era. More qualitative examples based on PaLM~2 are provided in Appendix \ref{appexdix:example}.

We compare the results of both approaches with \ourmodel{} which brings additional perspective to these results.
Leveraging the information about connections of entities across documents and documents themselves brings significant improvements up to 7 percentage points.
 
\section{Conclusions}\label{sec:conclusions}
\vspace{-1mm}
Our proposed model, \ourmodel{}, addresses limitations in existing ODQA methods by leveraging implicit connections between documents and strategically integrating AMR information. Our method identifies documents with valuable information significantly better even when this information is only weakly connected to the question context. This happens because documents connected by the document graph share information that is relevant for the final answer. 
We integrate key AMR information to improve performance without increasing computational cost. We also proposed two metrics to fairly evaluate the performance of a wide range of ranking scenarios including tied ranking scores. Furthermore, our investigation into the performance of PaLM~2 as a reranker emphasizes the significance of reranker model design in RAG, as even an advanced pre-trained LLM might face challenges in the reranking task.

Recently, papers such as \cite{ma2023zero,sun2023chatgpt} introduced methods for reranking listwise documents using LLMs. Despite this, our proposed metric MTRR remains valid for comparison with their approaches measured by MRR (mentioned in paper \cite{ma2023zero}). Thus our method has potential for broader adoption and comparison with existing approaches. Additionally, we're enthusiastic about investigating more advanced techniques to efficiently resolve ties in ranking scores produced by LLMs.
There are more directions for future study. For instance, designing more sophisticated models to better process AMR information and integrating this information into node \& edge features will bring further improvements in reranking.
Further, while a pre-trained LLM does not have impressive performance as a reranker itself, fine-tuning it may be extremely useful for enhancing the performance of RAG systems.

 \bibliographystyle{plainnat} \bibliography{bibliography}

\begin{thebibliography}{47}
\providecommand{\natexlab}[1]{#1}
\providecommand{\url}[1]{\texttt{#1}}
\expandafter\ifx\csname urlstyle\endcsname\relax
  \providecommand{\doi}[1]{doi: #1}\else
  \providecommand{\doi}{doi: \begingroup \urlstyle{rm}\Url}\fi

\bibitem[Asai et~al.(2020)Asai, Hashimoto, Hajishirzi, Socher, and
  Xiong]{pathretriever}
Akari Asai, Kazuma Hashimoto, Hannaneh Hajishirzi, Richard Socher, and Caiming
  Xiong.
\newblock Learning to retrieve reasoning paths over wikipedia graph for
  question answering.
\newblock In \emph{8th International Conference on Learning Representations,
  {ICLR} 2020, Addis Ababa, Ethiopia, April 26-30, 2020}. OpenReview.net, 2020.
\newblock URL \url{https://openreview.net/forum?id=SJgVHkrYDH}.

\bibitem[Bai et~al.(2021)Bai, Chen, Song, and Zhang]{amr4dial}
Xuefeng Bai, Yulong Chen, Linfeng Song, and Yue Zhang.
\newblock Semantic representation for dialogue modeling.
\newblock In \emph{Proceedings of the 59th Annual Meeting of the Association
  for Computational Linguistics and the 11th International Joint Conference on
  Natural Language Processing (Volume 1: Long Papers)}, pages 4430--4445,
  Online, August 2021. Association for Computational Linguistics.
\newblock \doi{10.18653/v1/2021.acl-long.342}.
\newblock URL \url{https://aclanthology.org/2021.acl-long.342}.

\bibitem[Bai et~al.(2022)Bai, Chen, and Zhang]{AMRBART}
Xuefeng Bai, Yulong Chen, and Yue Zhang.
\newblock Graph pre-training for {AMR} parsing and generation.
\newblock In \emph{Proceedings of the 60th Annual Meeting of the Association
  for Computational Linguistics (Volume 1: Long Papers)}, pages 6001--6015,
  Dublin, Ireland, May 2022. Association for Computational Linguistics.
\newblock URL \url{https://aclanthology.org/2022.acl-long.415}.

\bibitem[Banarescu et~al.(2013)Banarescu, Bonial, Cai, Georgescu, Griffitt,
  Hermjakob, Knight, Koehn, Palmer, and Schneider]{AMR}
Laura Banarescu, Claire Bonial, Shu Cai, Madalina Georgescu, Kira Griffitt, Ulf
  Hermjakob, Kevin Knight, Philipp Koehn, Martha Palmer, and Nathan Schneider.
\newblock {A}bstract {M}eaning {R}epresentation for sembanking.
\newblock In \emph{Proceedings of the 7th Linguistic Annotation Workshop and
  Interoperability with Discourse}, pages 178--186, Sofia, Bulgaria, August
  2013. Association for Computational Linguistics.
\newblock URL \url{https://aclanthology.org/W13-2322}.

\bibitem[Costa and Kulkarni(2018)]{costa2018leveraging}
Jose~Ortiz Costa and Anagha Kulkarni.
\newblock Leveraging knowledge graph for open-domain question answering.
\newblock In \emph{2018 IEEE/WIC/ACM International Conference on Web
  Intelligence (WI)}, pages 389--394. IEEE, 2018.

\bibitem[Devlin et~al.(2019)Devlin, Chang, Lee, and Toutanova]{BERT}
Jacob Devlin, Ming-Wei Chang, Kenton Lee, and Kristina Toutanova.
\newblock {BERT}: Pre-training of deep bidirectional transformers for language
  understanding.
\newblock In \emph{Proceedings of the 2019 Conference of the North {A}merican
  Chapter of the Association for Computational Linguistics: Human Language
  Technologies, Volume 1 (Long and Short Papers)}, pages 4171--4186,
  Minneapolis, Minnesota, June 2019. Association for Computational Linguistics.
\newblock \doi{10.18653/v1/N19-1423}.
\newblock URL \url{https://www.aclweb.org/anthology/N19-1423}.

\bibitem[Glass et~al.(2022)Glass, Rossiello, Chowdhury, Naik, Cai, and
  Gliozzo]{Re2G}
Michael Glass, Gaetano Rossiello, Md~Faisal~Mahbub Chowdhury, Ankita Naik,
  Pengshan Cai, and Alfio Gliozzo.
\newblock {R}e2{G}: Retrieve, rerank, generate.
\newblock In \emph{Proceedings of the 2022 Conference of the North American
  Chapter of the Association for Computational Linguistics: Human Language
  Technologies}, pages 2701--2715, Seattle, United States, July 2022.
  Association for Computational Linguistics.
\newblock \doi{10.18653/v1/2022.naacl-main.194}.
\newblock URL \url{https://aclanthology.org/2022.naacl-main.194}.

\bibitem[Gong and Cheng(2019)]{gong2019exploiting}
Liyu Gong and Qiang Cheng.
\newblock Exploiting edge features for graph neural networks.
\newblock In \emph{Proceedings of the IEEE/CVF conference on computer vision
  and pattern recognition}, pages 9211--9219, 2019.

\bibitem[Google et~al.(2023)Google, Anil, Dai, Firat, Johnson, Lepikhin,
  Passos, Shakeri, Taropa, Bailey, Chen, Chu, Clark, Shafey, Huang,
  Meier-Hellstern, Mishra, Moreira, Omernick, Robinson, Ruder, Tay, Xiao, Xu,
  Zhang, Abrego, Ahn, Austin, Barham, Botha, Bradbury, Brahma, Brooks, Catasta,
  Cheng, Cherry, Choquette-Choo, Chowdhery, Crepy, Dave, Dehghani, Dev, Devlin,
  Díaz, Du, Dyer, Feinberg, Feng, Fienber, Freitag, Garcia, Gehrmann,
  Gonzalez, Gur-Ari, Hand, Hashemi, Hou, Howland, Hu, Hui, Hurwitz, Isard,
  Ittycheriah, Jagielski, Jia, Kenealy, Krikun, Kudugunta, Lan, Lee, Lee, Li,
  Li, Li, Li, Li, Lim, Lin, Liu, Liu, Maggioni, Mahendru, Maynez, Misra,
  Moussalem, Nado, Nham, Ni, Nystrom, Parrish, Pellat, Polacek, Polozov, Pope,
  Qiao, Reif, Richter, Riley, Ros, Roy, Saeta, Samuel, Shelby, Slone, Smilkov,
  So, Sohn, Tokumine, Valter, Vasudevan, Vodrahalli, Wang, Wang, Wang, Wang,
  Wieting, Wu, Xu, Xu, Xue, Yin, Yu, Zhang, Zheng, Zheng, Zhou, Zhou, Petrov,
  and Wu]{anil2023palm}
Google, Rohan Anil, Andrew~M. Dai, Orhan Firat, Melvin Johnson, Dmitry
  Lepikhin, Alexandre Passos, Siamak Shakeri, Emanuel Taropa, Paige Bailey,
  Zhifeng Chen, Eric Chu, Jonathan~H. Clark, Laurent~El Shafey, Yanping Huang,
  Kathy Meier-Hellstern, Gaurav Mishra, Erica Moreira, Mark Omernick, Kevin
  Robinson, Sebastian Ruder, Yi~Tay, Kefan Xiao, Yuanzhong Xu, Yujing Zhang,
  Gustavo~Hernandez Abrego, Junwhan Ahn, Jacob Austin, Paul Barham, Jan Botha,
  James Bradbury, Siddhartha Brahma, Kevin Brooks, Michele Catasta, Yong Cheng,
  Colin Cherry, Christopher~A. Choquette-Choo, Aakanksha Chowdhery, Clément
  Crepy, Shachi Dave, Mostafa Dehghani, Sunipa Dev, Jacob Devlin, Mark Díaz,
  Nan Du, Ethan Dyer, Vlad Feinberg, Fangxiaoyu Feng, Vlad Fienber, Markus
  Freitag, Xavier Garcia, Sebastian Gehrmann, Lucas Gonzalez, Guy Gur-Ari,
  Steven Hand, Hadi Hashemi, Le~Hou, Joshua Howland, Andrea Hu, Jeffrey Hui,
  Jeremy Hurwitz, Michael Isard, Abe Ittycheriah, Matthew Jagielski, Wenhao
  Jia, Kathleen Kenealy, Maxim Krikun, Sneha Kudugunta, Chang Lan, Katherine
  Lee, Benjamin Lee, Eric Li, Music Li, Wei Li, YaGuang Li, Jian Li, Hyeontaek
  Lim, Hanzhao Lin, Zhongtao Liu, Frederick Liu, Marcello Maggioni, Aroma
  Mahendru, Joshua Maynez, Vedant Misra, Maysam Moussalem, Zachary Nado, John
  Nham, Eric Ni, Andrew Nystrom, Alicia Parrish, Marie Pellat, Martin Polacek,
  Alex Polozov, Reiner Pope, Siyuan Qiao, Emily Reif, Bryan Richter, Parker
  Riley, Alex~Castro Ros, Aurko Roy, Brennan Saeta, Rajkumar Samuel, Renee
  Shelby, Ambrose Slone, Daniel Smilkov, David~R. So, Daniel Sohn, Simon
  Tokumine, Dasha Valter, Vijay Vasudevan, Kiran Vodrahalli, Xuezhi Wang,
  Pidong Wang, Zirui Wang, Tao Wang, John Wieting, Yuhuai Wu, Kelvin Xu, Yunhan
  Xu, Linting Xue, Pengcheng Yin, Jiahui Yu, Qiao Zhang, Steven Zheng,
  Ce~Zheng, Weikang Zhou, Denny Zhou, Slav Petrov, and Yonghui Wu.
\newblock Palm 2 technical report, 2023.

\bibitem[Hamilton et~al.(2017)Hamilton, Ying, and
  Leskovec]{hamilton2017inductive}
Will Hamilton, Zhitao Ying, and Jure Leskovec.
\newblock Inductive representation learning on large graphs.
\newblock \emph{Advances in neural information processing systems}, 30, 2017.

\bibitem[Hou et~al.(2024)Hou, Zhang, Lin, Lu, Xie, McAuley, and
  Zhao]{hou2024large}
Yupeng Hou, Junjie Zhang, Zihan Lin, Hongyu Lu, Ruobing Xie, Julian McAuley,
  and Wayne~Xin Zhao.
\newblock Large language models are zero-shot rankers for recommender systems.
\newblock In \emph{European Conference on Information Retrieval}, pages
  364--381. Springer, 2024.

\bibitem[Huang et~al.(2023)Huang, Wei, Yue, Zhao, Chen, Li, Jiang, Chang,
  Zhang, Zhang, et~al.]{huang2023dsqa}
Dengrong Huang, Zizhong Wei, Aizhen Yue, Xuan Zhao, Zhaoliang Chen, Rui Li, Kai
  Jiang, Bingxin Chang, Qilai Zhang, Sijia Zhang, et~al.
\newblock Dsqa-llm: Domain-specific intelligent question answering based on
  large language model.
\newblock In \emph{International Conference on AI-generated Content}, pages
  170--180. Springer, 2023.

\bibitem[Izacard and Grave(2021)]{FiD}
Gautier Izacard and {\'E}douard Grave.
\newblock Leveraging passage retrieval with generative models for open domain
  question answering.
\newblock In \emph{Proceedings of the 16th Conference of the European Chapter
  of the Association for Computational Linguistics: Main Volume}, pages
  874--880, 2021.

\bibitem[Joshi et~al.(2017)Joshi, Choi, Weld, and
  Zettlemoyer]{joshi-etal-2017-triviaqa}
Mandar Joshi, Eunsol Choi, Daniel Weld, and Luke Zettlemoyer.
\newblock {T}rivia{QA}: A large scale distantly supervised challenge dataset
  for reading comprehension.
\newblock In Regina Barzilay and Min-Yen Kan, editors, \emph{Proceedings of the
  55th Annual Meeting of the Association for Computational Linguistics (Volume
  1: Long Papers)}, pages 1601--1611, Vancouver, Canada, July 2017. Association
  for Computational Linguistics.
\newblock \doi{10.18653/v1/P17-1147}.
\newblock URL \url{https://aclanthology.org/P17-1147}.

\bibitem[Ju et~al.(2022)Ju, Yu, Zhao, Zhang, and Ye]{GRAPE}
Mingxuan Ju, Wenhao Yu, Tong Zhao, Chuxu Zhang, and Yanfang Ye.
\newblock Grape: Knowledge graph enhanced passage reader for open-domain
  question answering.
\newblock In \emph{Findings of Empirical Methods in Natural Language
  Processing}, 2022.

\bibitem[Karpukhin et~al.(2020)Karpukhin, Oguz, Min, Lewis, Wu, Edunov, Chen,
  and Yih]{DPR}
Vladimir Karpukhin, Barlas Oguz, Sewon Min, Patrick Lewis, Ledell Wu, Sergey
  Edunov, Danqi Chen, and Wen-tau Yih.
\newblock Dense passage retrieval for open-domain question answering.
\newblock In \emph{Proceedings of the 2020 Conference on Empirical Methods in
  Natural Language Processing (EMNLP)}, pages 6769--6781, Online, 2020.
  Association for Computational Linguistics.
\newblock \doi{10.18653/v1/2020.emnlp-main.550}.
\newblock URL \url{https://aclanthology.org/2020.emnlp-main.550}.

\bibitem[Kipf and Welling(2017)]{kipf2017semi}
Thomas~N Kipf and Max Welling.
\newblock Semi-supervised classification with graph convolutional networks.
\newblock In \emph{ICLR}, 2017.

\bibitem[Kwiatkowski et~al.(2019)Kwiatkowski, Palomaki, Redfield, Collins,
  Parikh, Alberti, Epstein, Polosukhin, Devlin, Lee,
  et~al.]{kwiatkowski2019natural}
Tom Kwiatkowski, Jennimaria Palomaki, Olivia Redfield, Michael Collins, Ankur
  Parikh, Chris Alberti, Danielle Epstein, Illia Polosukhin, Jacob Devlin,
  Kenton Lee, et~al.
\newblock Natural questions: a benchmark for question answering research.
\newblock \emph{Transactions of the Association for Computational Linguistics},
  7:\penalty0 453--466, 2019.

\bibitem[Lewis et~al.(2020{\natexlab{a}})Lewis, Liu, Goyal, Ghazvininejad,
  Mohamed, Levy, Stoyanov, and Zettlemoyer]{bart}
Mike Lewis, Yinhan Liu, Naman Goyal, Marjan Ghazvininejad, Abdelrahman Mohamed,
  Omer Levy, Veselin Stoyanov, and Luke Zettlemoyer.
\newblock {BART}: Denoising sequence-to-sequence pre-training for natural
  language generation, translation, and comprehension.
\newblock In \emph{Proceedings of the 58th Annual Meeting of the Association
  for Computational Linguistics}, pages 7871--7880, Online, July
  2020{\natexlab{a}}. Association for Computational Linguistics.
\newblock \doi{10.18653/v1/2020.acl-main.703}.
\newblock URL \url{https://www.aclweb.org/anthology/2020.acl-main.703}.

\bibitem[Lewis et~al.(2020{\natexlab{b}})Lewis, Perez, Piktus, Petroni,
  Karpukhin, Goyal, K{\"u}ttler, Lewis, Yih, Rockt{\"a}schel,
  et~al.]{lewis2020retrieval}
Patrick Lewis, Ethan Perez, Aleksandra Piktus, Fabio Petroni, Vladimir
  Karpukhin, Naman Goyal, Heinrich K{\"u}ttler, Mike Lewis, Wen-tau Yih, Tim
  Rockt{\"a}schel, et~al.
\newblock Retrieval-augmented generation for knowledge-intensive nlp tasks.
\newblock \emph{NeurIPS}, 33:\penalty0 9459--9474, 2020{\natexlab{b}}.

\bibitem[Li et~al.(2017)Li, Song, and Luo]{li2017improving}
Yuncheng Li, Yale Song, and Jiebo Luo.
\newblock Improving pairwise ranking for multi-label image classification.
\newblock In \emph{Proceedings of the IEEE conference on computer vision and
  pattern recognition}, pages 3617--3625, 2017.

\bibitem[Li et~al.(2023)Li, Zhang, Zhang, Long, Xie, and Zhang]{li2023general}
Zehan Li, Xin Zhang, Yanzhao Zhang, Dingkun Long, Pengjun Xie, and Meishan
  Zhang.
\newblock Towards general text embeddings with multi-stage contrastive
  learning, 2023.

\bibitem[Liu et~al.(2019)Liu, Ott, Goyal, Du, Joshi, Chen, Levy, Lewis,
  Zettlemoyer, and Stoyanov]{RoBERTa}
Yinhan Liu, Myle Ott, Naman Goyal, Jingfei Du, Mandar Joshi, Danqi Chen, Omer
  Levy, Mike Lewis, Luke Zettlemoyer, and Veselin Stoyanov.
\newblock Roberta: A robustly optimized bert pretraining approach.
\newblock \emph{arXiv preprint arXiv:1907.11692}, 2019.

\bibitem[Liu and Shao(2022)]{liu2022retromae}
Zheng Liu and Yingxia Shao.
\newblock Retro{MAE}: Pre-training retrieval-oriented transformers via masked
  auto-encoder.
\newblock \emph{arXiv preprint arXiv:2205.12035}, 2022.

\bibitem[Loshchilov and Hutter(2019)]{adamw}
Ilya Loshchilov and Frank Hutter.
\newblock Decoupled weight decay regularization.
\newblock In \emph{7th International Conference on Learning Representations,
  {ICLR} 2019, New Orleans, LA, USA, May 6-9, 2019}. OpenReview.net, 2019.
\newblock URL \url{https://openreview.net/forum?id=Bkg6RiCqY7}.

\bibitem[Ma et~al.(2023{\natexlab{a}})Ma, Wang, Yang, Wei, and Lin]{ma2023fine}
Xueguang Ma, Liang Wang, Nan Yang, Furu Wei, and Jimmy Lin.
\newblock Fine-tuning llama for multi-stage text retrieval.
\newblock \emph{arXiv preprint arXiv:2310.08319}, 2023{\natexlab{a}}.

\bibitem[Ma et~al.(2023{\natexlab{b}})Ma, Zhang, Pradeep, and Lin]{ma2023zero}
Xueguang Ma, Xinyu Zhang, Ronak Pradeep, and Jimmy Lin.
\newblock Zero-shot listwise document reranking with a large language model.
\newblock \emph{arXiv preprint arXiv:2305.02156}, 2023{\natexlab{b}}.

\bibitem[Ma et~al.(2023{\natexlab{c}})Ma, Cao, Hong, and Sun]{ma2023large}
Yubo Ma, Yixin Cao, Yong Hong, and Aixin Sun.
\newblock Large language model is not a good few-shot information extractor,
  but a good reranker for hard samples!
\newblock In \emph{Findings of the Association for Computational Linguistics:
  EMNLP 2023}, pages 10572--10601, 2023{\natexlab{c}}.

\bibitem[Naseem et~al.(2021)Naseem, Blodgett, Kumaravel, O'Gorman, Lee,
  Flanigan, Astudillo, Florian, Roukos, and Schneider]{DocAMR}
Tahira Naseem, Austin Blodgett, Sadhana Kumaravel, Timothy~J. O'Gorman,
  Young-Suk Lee, Jeffrey Flanigan, Ram{\'o}n~Fern{\'a}ndez Astudillo, Radu
  Florian, Salim Roukos, and Nathan Schneider.
\newblock Docamr: Multi-sentence amr representation and evaluation.
\newblock In \emph{North American Chapter of the Association for Computational
  Linguistics}, 2021.

\bibitem[Nogueira et~al.(2020)Nogueira, Jiang, Pradeep, and
  Lin]{nogueira2020document}
Rodrigo Nogueira, Zhiying Jiang, Ronak Pradeep, and Jimmy Lin.
\newblock Document ranking with a pretrained sequence-to-sequence model.
\newblock In \emph{Findings of the Association for Computational Linguistics:
  EMNLP 2020}, pages 708--718, 2020.

\bibitem[OpenAI({\natexlab{a}})]{Chatgpt}
OpenAI.
\newblock Chat{GPT}.
\newblock \url{https://openai.com/ research/chatgpt.}, {\natexlab{a}}.

\bibitem[OpenAI({\natexlab{b}})]{gpt4}
OpenAI.
\newblock {GPT}-4.
\newblock \url{https://openai.com/gpt-4}, {\natexlab{b}}.

\bibitem[Park et~al.(2023)Park, Lee, Seo, Kim, Kang, and Na]{park2023rink}
Eunhwan Park, Sung-Min Lee, Dearyong Seo, Seonhoon Kim, Inho Kang, and
  Seung-Hoon Na.
\newblock Rink: reader-inherited evidence reranker for table-and-text open
  domain question answering.
\newblock In \emph{Proceedings of the AAAI Conference on Artificial
  Intelligence}, volume~37, pages 13446--13456, 2023.

\bibitem[Pradeep et~al.(2022)Pradeep, Liu, Zhang, Li, Yates, and
  Lin]{pradeep2022squeezing}
Ronak Pradeep, Yuqi Liu, Xinyu Zhang, Yilin Li, Andrew Yates, and Jimmy Lin.
\newblock Squeezing water from a stone: a bag of tricks for further improving
  cross-encoder effectiveness for reranking.
\newblock In \emph{European Conference on Information Retrieval}, pages
  655--670. Springer, 2022.

\bibitem[Siriwardhana et~al.(2023)Siriwardhana, Weerasekera, Wen, Kaluarachchi,
  Rana, and Nanayakkara]{siriwardhana2023improving}
Shamane Siriwardhana, Rivindu Weerasekera, Elliott Wen, Tharindu Kaluarachchi,
  Rajib Rana, and Suranga Nanayakkara.
\newblock Improving the domain adaptation of retrieval augmented generation
  (rag) models for open domain question answering.
\newblock \emph{Transactions of the Association for Computational Linguistics},
  11:\penalty0 1--17, 2023.

\bibitem[Sun et~al.(2023)Sun, Yan, Ma, Wang, Ren, Chen, Yin, and
  Ren]{sun2023chatgpt}
Weiwei Sun, Lingyong Yan, Xinyu Ma, Shuaiqiang Wang, Pengjie Ren, Zhumin Chen,
  Dawei Yin, and Zhaochun Ren.
\newblock Is chatgpt good at search? investigating large language models as
  re-ranking agents.
\newblock In \emph{Proceedings of the 2023 Conference on Empirical Methods in
  Natural Language Processing}, pages 14918--14937, 2023.

\bibitem[Tan et~al.(2023)Tan, Min, Li, Li, Hu, Chen, and Qi]{tan2023can}
Yiming Tan, Dehai Min, Yu~Li, Wenbo Li, Nan Hu, Yongrui Chen, and Guilin Qi.
\newblock Can chatgpt replace traditional kbqa models? an in-depth analysis of
  the question answering performance of the gpt llm family.
\newblock In \emph{International Semantic Web Conference}, pages 348--367.
  Springer, 2023.

\bibitem[Touvron et~al.(2023)Touvron, Lavril, Izacard, Martinet, Lachaux,
  Lacroix, Rozi{\`e}re, Goyal, Hambro, Azhar, et~al.]{touvron2023llama}
Hugo Touvron, Thibaut Lavril, Gautier Izacard, Xavier Martinet, Marie-Anne
  Lachaux, Timoth{\'e}e Lacroix, Baptiste Rozi{\`e}re, Naman Goyal, Eric
  Hambro, Faisal Azhar, et~al.
\newblock Llama: Open and efficient foundation language models.
\newblock \emph{arXiv preprint arXiv:2302.13971}, 2023.

\bibitem[Veli{\v{c}}kovi{\'c} et~al.(2018)Veli{\v{c}}kovi{\'c}, Cucurull,
  Casanova, Romero, Li{\`o}, and Bengio]{velivckovic2018graph}
Petar Veli{\v{c}}kovi{\'c}, Guillem Cucurull, Arantxa Casanova, Adriana Romero,
  Pietro Li{\`o}, and Yoshua Bengio.
\newblock Graph attention networks.
\newblock In \emph{International Conference on Learning Representations}, 2018.

\bibitem[Voorhees and Tice(2000)]{voorhees-tice-2000-trec}
Ellen~M. Voorhees and Dawn~M. Tice.
\newblock The {TREC}-8 question answering track.
\newblock In M.~Gavrilidou, G.~Carayannis, S.~Markantonatou, S.~Piperidis, and
  G.~Stainhauer, editors, \emph{Proceedings of the Second International
  Conference on Language Resources and Evaluation ({LREC}{'}00)}, Athens,
  Greece, May 2000. European Language Resources Association (ELRA).
\newblock URL \url{http://www.lrec-conf.org/proceedings/lrec2000/pdf/26.pdf}.

\bibitem[Wang et~al.(2023{\natexlab{a}})Wang, Cheng, Xu, Ding, Wang, and
  Zhang]{wang2023evaluating}
Cunxiang Wang, Sirui Cheng, Zhikun Xu, Bowen Ding, Yidong Wang, and Yue Zhang.
\newblock Evaluating open question answering evaluation.
\newblock \emph{arXiv preprint arXiv:2305.12421}, 2023{\natexlab{a}}.

\bibitem[Wang et~al.(2023{\natexlab{b}})Wang, Xu, Guo, Hu, Bai, Zhang, and
  Zhang]{wang-etal-2023-exploiting}
Cunxiang Wang, Zhikun Xu, Qipeng Guo, Xiangkun Hu, Xuefeng Bai, Zheng Zhang,
  and Yue Zhang.
\newblock Exploiting {A}bstract {M}eaning {R}epresentation for open-domain
  question answering.
\newblock In Anna Rogers, Jordan Boyd-Graber, and Naoaki Okazaki, editors,
  \emph{Findings of the Association for Computational Linguistics: ACL 2023},
  pages 2083--2096, Toronto, Canada, July 2023{\natexlab{b}}. Association for
  Computational Linguistics.
\newblock \doi{10.18653/v1/2023.findings-acl.131}.
\newblock URL \url{https://aclanthology.org/2023.findings-acl.131}.

\bibitem[Wolf et~al.(2019)Wolf, Debut, Sanh, Chaumond, Delangue, Moi, Cistac,
  Rault, Louf, Funtowicz, et~al.]{wolf2019huggingface}
Thomas Wolf, Lysandre Debut, Victor Sanh, Julien Chaumond, Clement Delangue,
  Anthony Moi, Pierric Cistac, Tim Rault, R{\'e}mi Louf, Morgan Funtowicz,
  et~al.
\newblock Huggingface's transformers: State-of-the-art natural language
  processing.
\newblock \emph{arXiv preprint arXiv:1910.03771}, 2019.
\newblock URL \url{https://arxiv.org/abs/1910.03771}.

\bibitem[Xiao et~al.(2023)Xiao, Liu, Zhang, and Muennighoff]{bge_embedding}
Shitao Xiao, Zheng Liu, Peitian Zhang, and Niklas Muennighoff.
\newblock C-pack: Packaged resources to advance general chinese embedding,
  2023.

\bibitem[Xu et~al.(2018)Xu, Hu, Leskovec, and Jegelka]{xu2018powerful}
Keyulu Xu, Weihua Hu, Jure Leskovec, and Stefanie Jegelka.
\newblock How powerful are graph neural networks?
\newblock In \emph{International Conference on Learning Representations}, 2018.

\bibitem[Yu et~al.(2022)Yu, Zhu, Fang, Yu, Wang, Xu, Ren, Yang, and
  Zeng]{KG-FiD}
Donghan Yu, Chenguang Zhu, Yuwei Fang, Wenhao Yu, Shuohang Wang, Yichong Xu,
  Xiang Ren, Yiming Yang, and Michael Zeng.
\newblock {KG}-{F}i{D}: Infusing knowledge graph in fusion-in-decoder for
  open-domain question answering.
\newblock In \emph{ACL}, pages 4961--4974, Dublin, Ireland, May 2022.
  Association for Computational Linguistics.
\newblock \doi{10.18653/v1/2022.acl-long.340}.
\newblock URL \url{https://aclanthology.org/2022.acl-long.340}.

\bibitem[Zhuang et~al.(2023)Zhuang, Qin, Jagerman, Hui, Ma, Lu, Ni, Wang, and
  Bendersky]{zhuang2023rankt5}
Honglei Zhuang, Zhen Qin, Rolf Jagerman, Kai Hui, Ji~Ma, Jing Lu, Jianmo Ni,
  Xuanhui Wang, and Michael Bendersky.
\newblock Rankt5: Fine-tuning t5 for text ranking with ranking losses.
\newblock In \emph{SIGIR}, pages 2308--2313, 2023.

\end{thebibliography}
\appendix
\section{Dataset Statistics}\label{appexdix:data}
\begin{figure}[h]
\begin{tcbitemize}[raster equal height=rows,
raster columns=4, raster halign=center,
raster every box/.style=blankest]
\mysubfig{NQ-train}{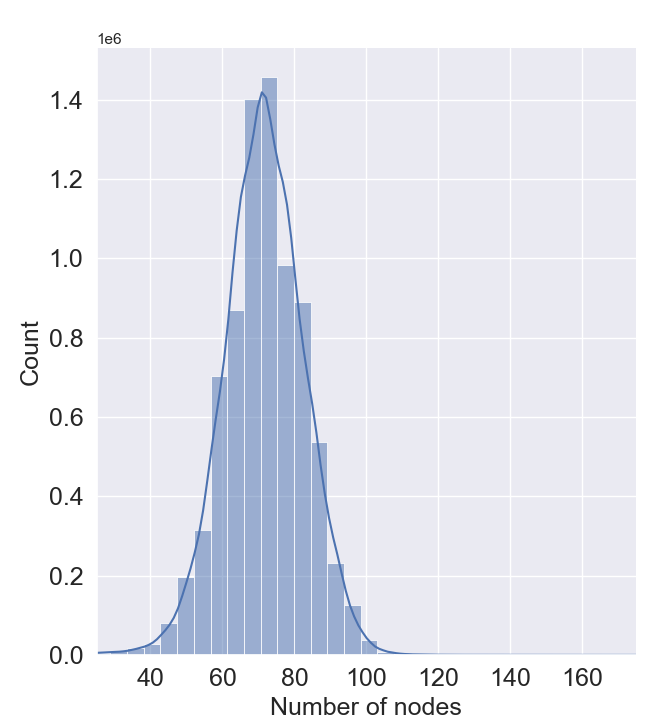}
\mysubfig{NQ-train}{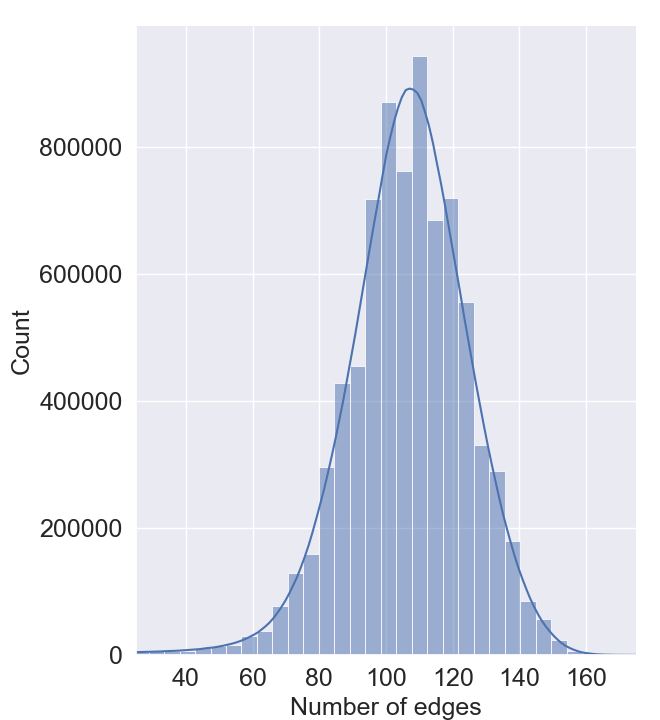}
\mysubfig{NQ-dev}{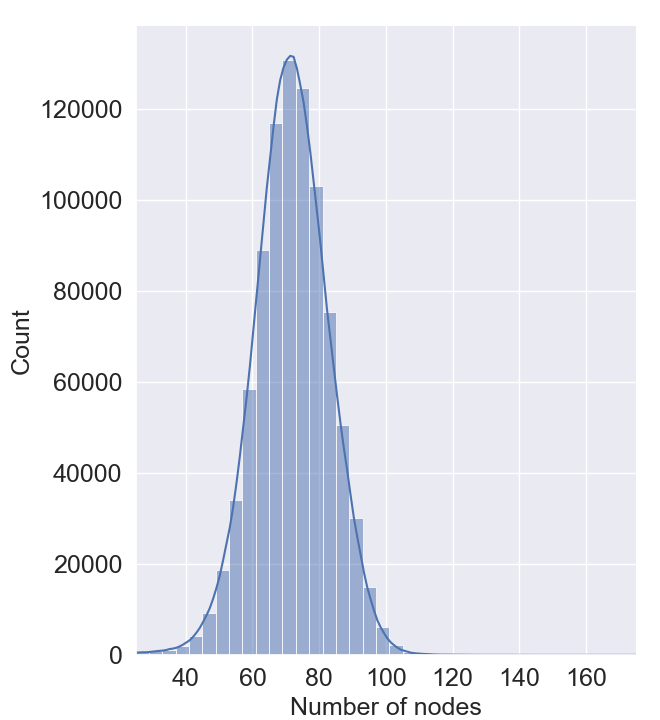}
\mysubfig{NQ-dev}{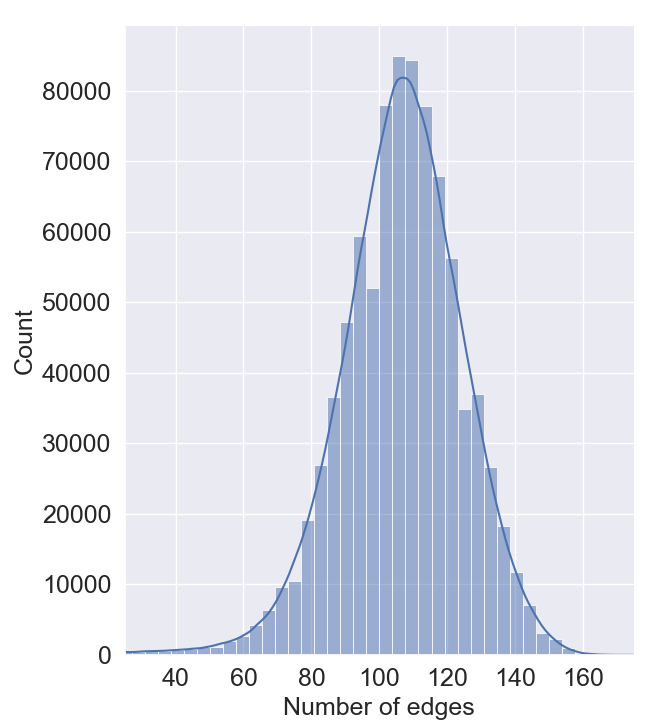}
\mysubfig{NQ-test}{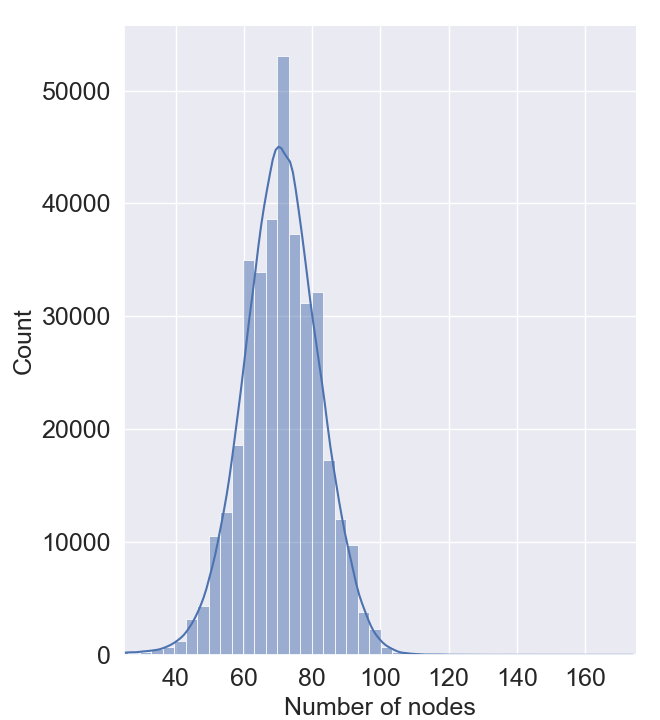}
\mysubfig{NQ-test}{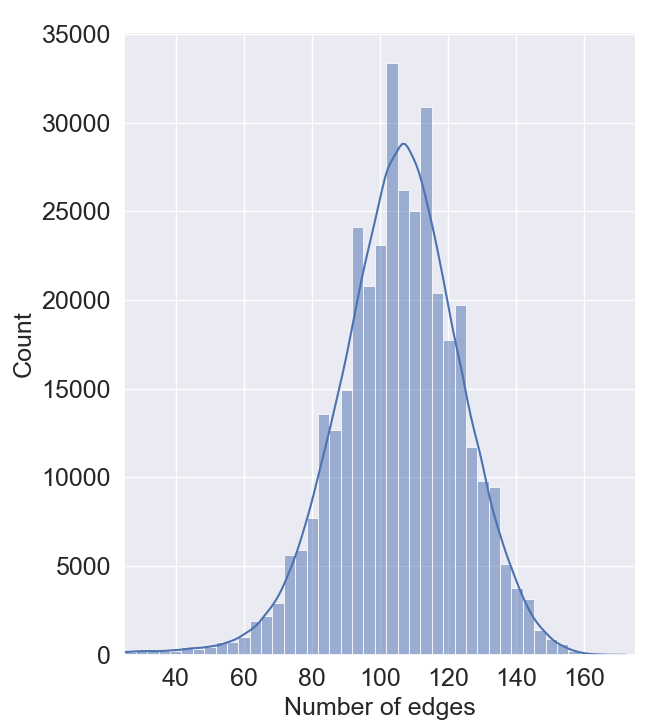}
\mysubfig{TQA-train}{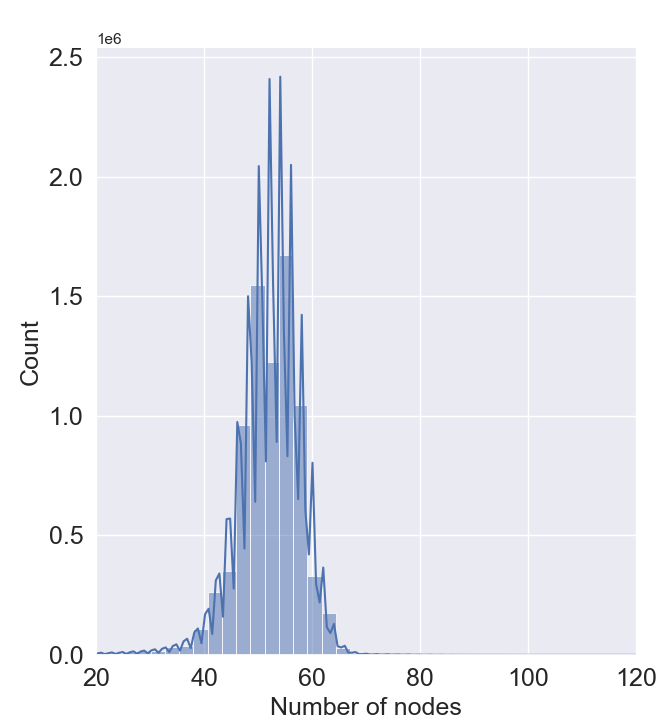}
\mysubfig{TQA-train}{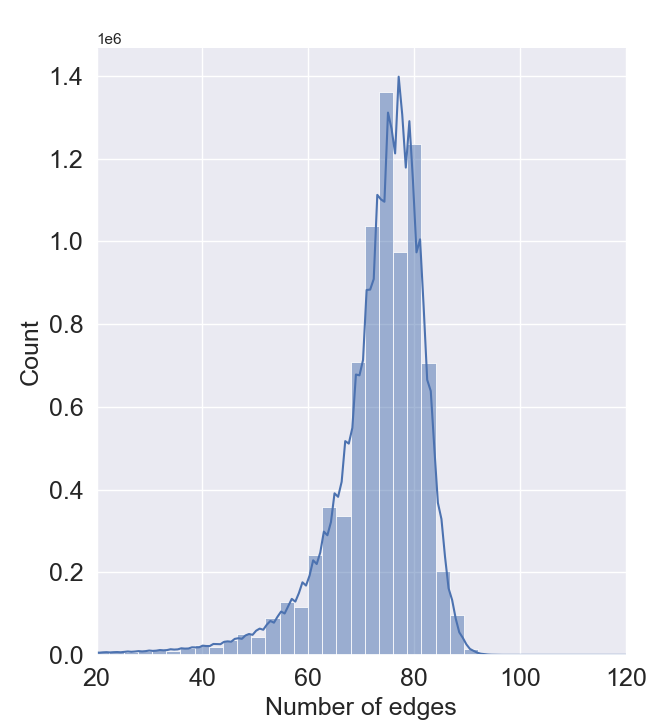}
\mysubfig{TQA-dev}{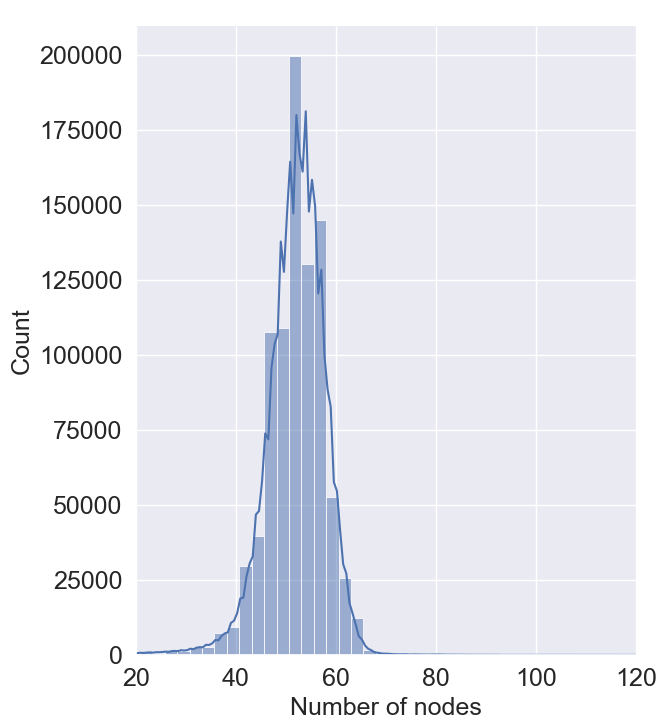}
\mysubfig{TQA-dev}{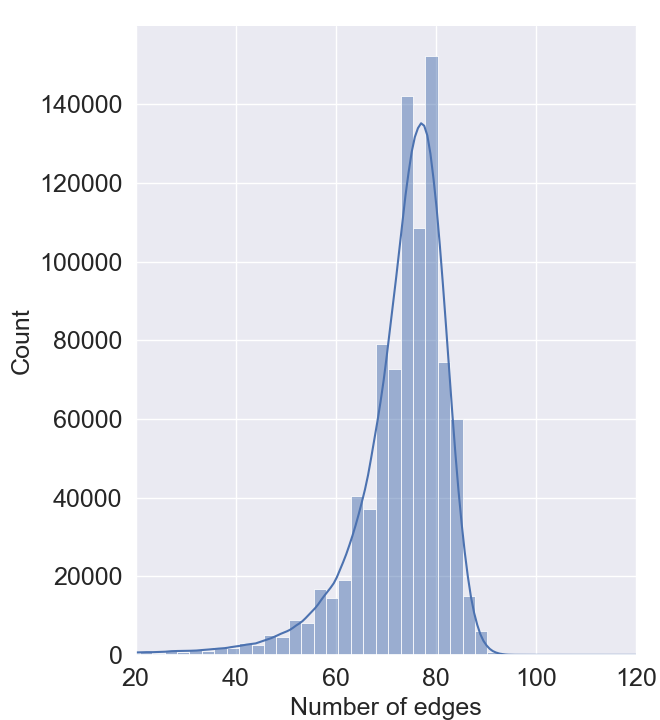}
\mysubfig{TQA-test}{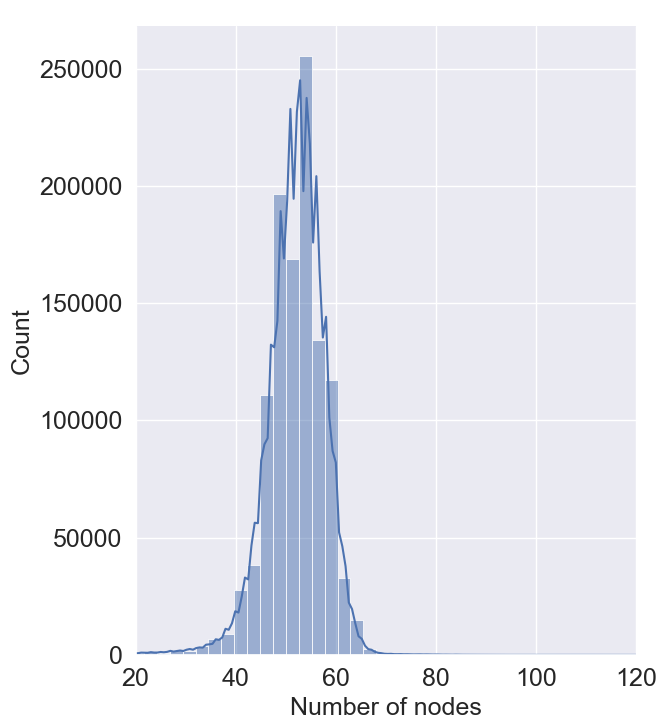}
\mysubfig{TQA-test}{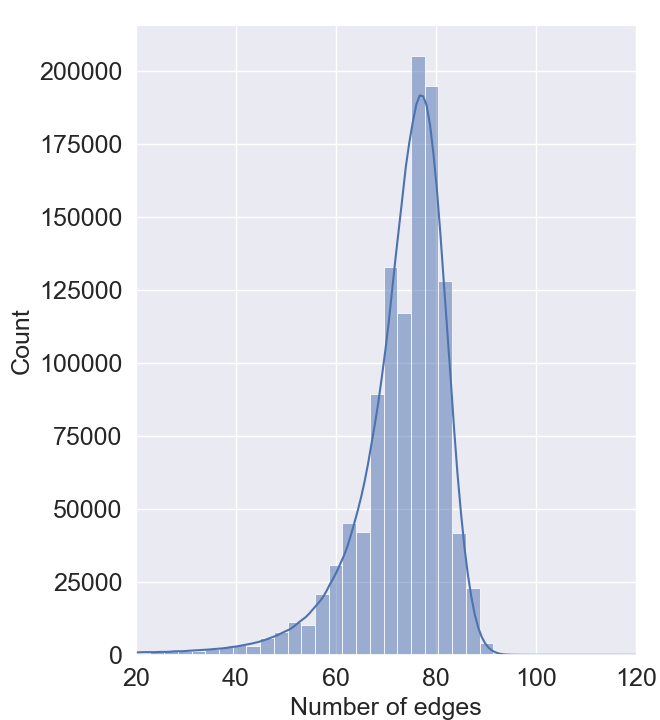}
\end{tcbitemize}
\caption{Number of nodes and edges in AMR graphs in train/dev/test set of dataset NQ and TQA.}\label{stat}
\end{figure}

\begin{figure}[t]
\begin{tcbitemize}[raster equal height=rows,
raster columns=4, raster halign=center,
raster every box/.style=blankest]
\mysubfig{NQ-Positive}{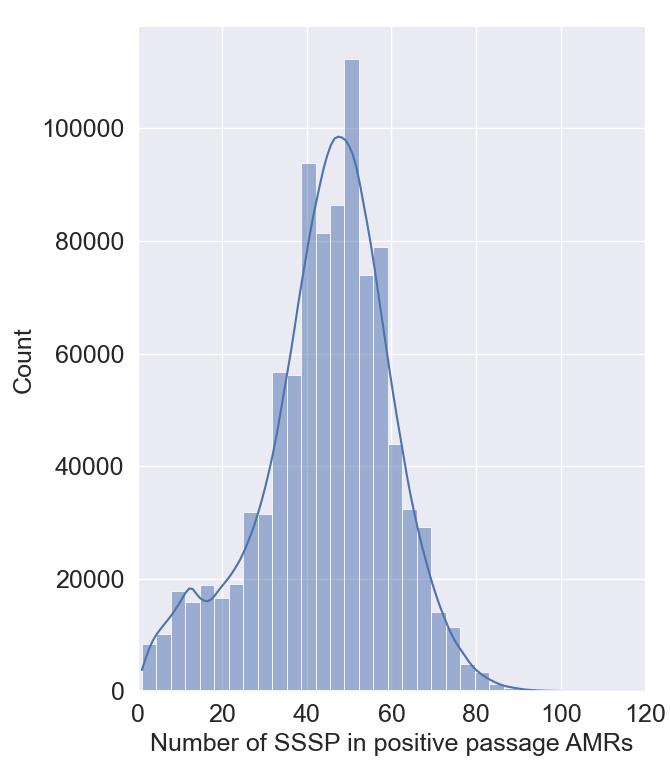}
\mysubfig{NQ-Negative}{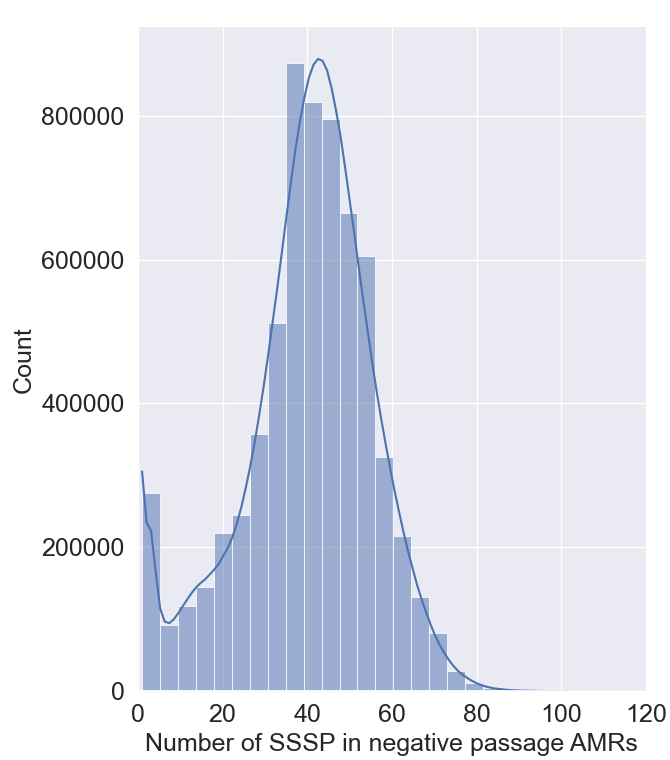}
\mysubfig{TQA-Positive}{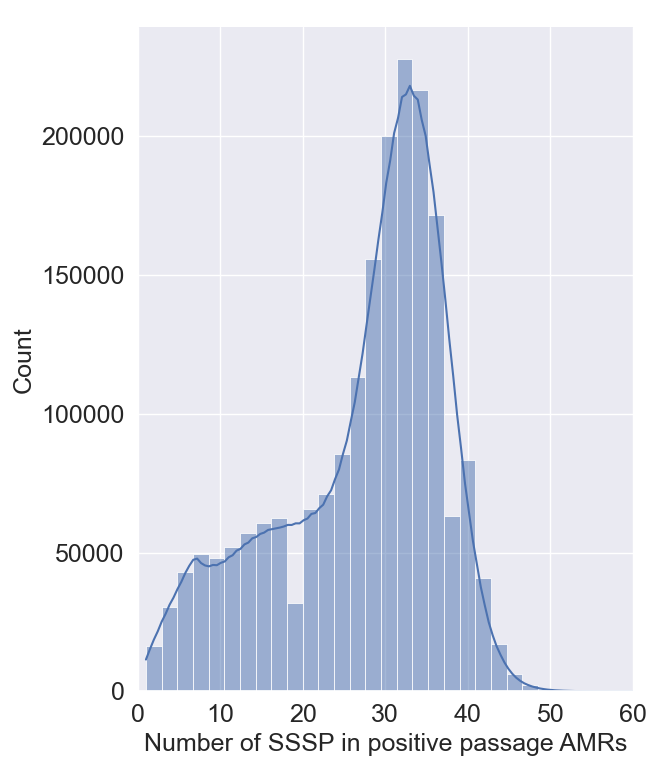}
\mysubfig{TQA-Negative}{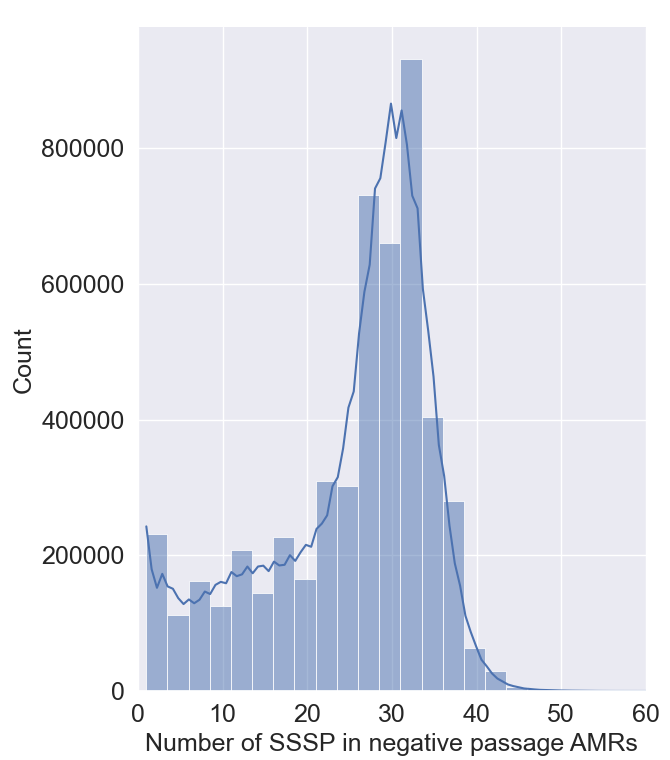}
\end{tcbitemize}
\caption{Number of SSSPs AMR graphs in train set of dataset NQ and TQA.}\label{fig:shortest path}
\end{figure}

\begin{table}[htp]
\centering
\begin{tabular}{|l|l|l|l|}
\hline
 & Train & Dev & Test \\ \hline
Natural Questions & 79168 & 8757 & 3610 \\ \hline
TriviaQA & 78785 & 8837 & 11313 \\ \hline
\end{tabular}
\caption{Dataset Statistics.}\label{table:dataset}
\end{table}
In Fig. \ref{stat}, we illustrate the AMR graph statistics in the datasets Natural Questions (NQ) and TriviaQA.
To better illustrate the structure of the shortest path, we also conduct some experiments to show the statistic of the shortest path in the AMR graph, see Fig \ref{fig:shortest path}. We analyze the shortest single source paths (SSSPs) in the AMR graphs of documents and try to establish the connection between question contexts and document contexts. The analysis reveals a notable trend in the AMR graphs of documents, indicating that certain negative documents cannot establish adequate connections to the question context within their text. 
This pattern brings insights into the encoding process to enhance reranking performance.

\section{Simulation Results with Different GNN Models.}\label{appendix:gnn}
Besides the GCN \citep{kipf2017semi} model considered in the main manuscript, we compare the simulation results with different GNN models in this section. Specifically, under the same setting as the GCN model in Ember (HPs-T) from Table \ref{table:embedding}, we use GAT \citep{velivckovic2018graph} with additional parameter number of heads being $8$, GraphSage \citep{hamilton2017inductive} with the aggregation choice being `lstm', and GIN \citep{xu2018powerful} with the aggregation choice being `mean'. The comparison results are illustrated in Table \ref{table:gnn}. For the convenience of comparison, we directly add two results from Section \ref{sec:exp_baseline}, i.e., BART-GST and GCN (i.e., Ember (HPs-T) in Table \ref{table:embedding}). It shows that the GCN model still outperforms in most cases. This may be due to the document graphs considered in our paper being very small, while the advanced GNN model usually targets handling thousands, or millions of nodes in the graph. Besides, our model has already taken the edge feature into consideration, which may lead to overfitting if introducing more weight parameters.

\begin{table}[htp]
\footnotesize
\centering
\newcolumntype{C}{>{\centering\arraybackslash}X}
\begin{tabularx}{\linewidth}{X|CC|CC|CC|CC}
\toprule
 & \multicolumn{4}{c|}{\textbf{NQ}} & \multicolumn{4}{c}{\textbf{TQA}} \\ 
\midrule
\begin{tabular}[c]{@{}l@{}}\textbf{Embedding}\\ \textbf{/Metric}\end{tabular} & \textbf{MRR\_dev} & \textbf{MRR\_test} & \textbf{MH\_dev} & \textbf{MH\_test} & \textbf{MRR\_dev} & \textbf{MRR\_test} & \textbf{MH\_dev} & \textbf{MH\_test} \\ 
\midrule
\mbox{BART-GST} & 28.4 & 25.0 & \textbf{53.2} & 48.7 & 17.5 & 17.6 & 39.1 & 39.5 \\ 
GCN & 28.9 & 27.7 & 51.1 & \textbf{50.0} & \textbf{20.0} & \textbf{19.4} & {41.6} & \textbf{41.4} \\ 
GAT & 28.1 & 27.1 & 52.3 & 47.2 & 19.1 & 18.9 & \textbf{43.0} & 41.0 \\ 
GraphSage & \textbf{29.8} & 26.5 & 52.3 & 47.2 & 19.6 & 18.4 & 42.9 & 39.7 \\ 
GIN & 28.4 & \textbf{27.8} & 50.2 & 48.5 & 19.7 & 18.9 & 42.2 & 39.3 \\
\bottomrule
\end{tabularx}
\vspace{2mm}
\caption{Results of \ourmodel{} with different GNN models. We use Mean Hits @ 10.}
\label{table:gnn}
\end{table}

\begin{figure}[tp]
    \centering
    \includegraphics[width=\textwidth]{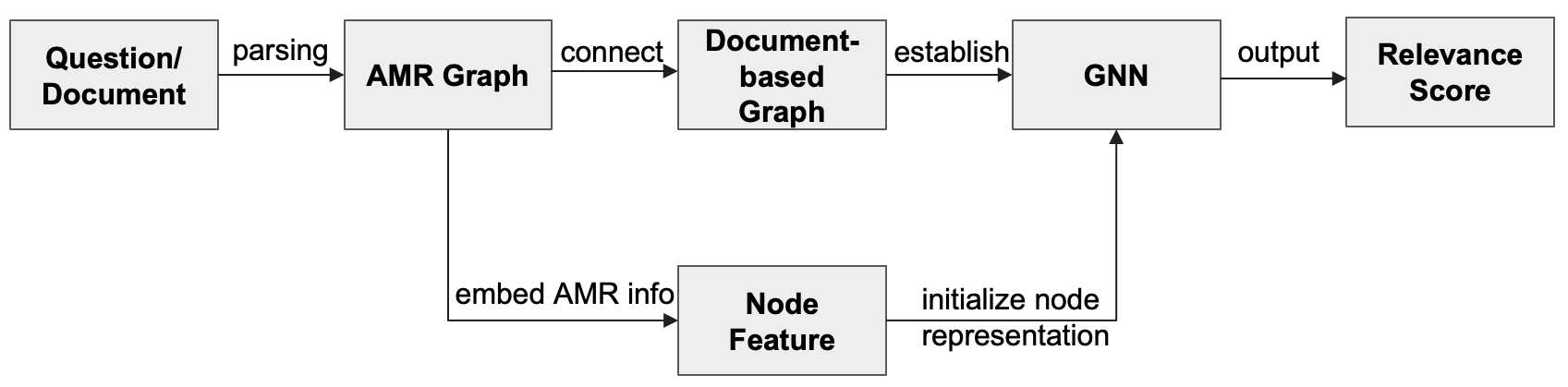}
    \caption{The pipeline of G-RAG.}
\label{fig:model_pipeline}
    \vspace*{-2mm}
\end{figure}

\section{Qualitative Examples}\label{appexdix:example}
We take the ranking scores given by palm~2 L as a baseline to investigate how the graph-based model brings benefits to reranking in Open-Domain Question Answering. Since TQA is a much more complex dataset with more positive documents, we take an example from TQA.
\begin{tcolorbox}
\textbf{Question:} Ol' Blue Eyes is the nickname of?\\
\textbf{Gold Answer:} [`Sinatra (film)',
 `Biography of Frank Sinatra',
 `Columbus Day Riot',
 `Life of Frank Sinatra',
 `A Voice in Time: 1939–1952',
 `Sinatra',
 `Biography of frank sinatra',
 `Ol’ Blue Eyes',
 `A Voice in Time: 1939-1952',
 `Political beliefs of frank sinatra',
 `Franck Sinatra',
 `Old Blue Eyes',
 `Frank Sinatra',
 `Frank Sinatra I',
 `Francis Albert Frank Sinatra',
 `Francis A. Sinatra',
 `Frank Sinatra, Sr.',
 `Francis Albert Sinatra',
 `Political beliefs of Frank Sinatra',
 `Old blue eyes',
 `Frank sanatra',
 `Frank sinatra',
 `Frank senatra',
 `FBI Files on Frank Sinatra',
 `Francis Sinatra']\\
\textbf{Number of Positive documents:} 24 positive documents out of 100 documents
 
\end{tcolorbox}
The following are Top-10 documents given by the proposed GNN-based reranker. Each document is accompanied by relevant information about its AMR graph, including the number of nodes and edges, as well as the count of single-source shortest paths (SSSPs) originating from the node labeled ``question". If the node ``question" is not present in the AMR graph, the SSSPs count is noted as 0. Additionally, we present the corresponding score assigned by palm~2-L and its rank based on the palm~2 reranker. The ranking assigned by the retriever DPR is also provided for reference. \citep{DPR}. 
\begin{tcolorbox}
\textbf{1st:} \hl{Sinatra} in 1998, for example, the building was bathed in blue light to represent the singer's \hl{nickname "Ol' Blue Eyes"}. After actress Fay Wray, who starred in "King Kong", died in September 2004, the building lights were extinguished for 15 minutes. The floodlights bathed the building in red, white, and blue for several months after the destruction of the World Trade Center in September 2001, then reverted to the standard schedule. On June 4, 2002, the Empire State Building donned purple and gold (the royal colors of Elizabeth II), in thanks for the United Kingdom playing the Star Spangled Banner\\
- - - - - - - - - - - - - - - - - - - - - - - - - - - - - - - - - - - - - - - - - - - - -  \\
\textbf{AMR graph information:} \# nodes: 51,  \# edges 82, \# SSSP: 32\\
\textbf{Score by palm~2: }50/100,  \textbf{Rank by palm~2: }9/100\\
\textbf{Rank by DPR:} 5/100
\end{tcolorbox}

\begin{tcolorbox}
\textbf{2nd:} and actively campaigned for presidents such as Harry S. Truman, John F. Kennedy and Ronald Reagan. In crime, the FBI investigated \hl{Sinatra} and his alleged relationship with the Mafia. While \hl{Sinatra} never learned how to read music, he had an impressive understanding of it, and he worked very hard from a young age to improve his abilities in all aspects of music. A perfectionist, renowned for his dress sense and performing presence, he always insisted on recording live with his band. His bright blue eyes earned him the popular \hl{nickname "Ol' Blue Eyes"}. Sinatra led a colorful personal life, and\\
- - - - - - - - - - - - - - - - - - - - - - - - - - - - - - - - - - - - - - - - - - - - -  \\
\textbf{AMR graph information:} \# nodes: 53,  \# edges 75, \# SSSP: 34\\
\textbf{Score by palm~2: }100/100,  \textbf{Rank by palm~2: }1/100\\
\textbf{Rank by DPR:} 1/100
\end{tcolorbox}

\begin{tcolorbox}
\textbf{3rd:} claimed that \hl{Sinatra} had grown "tired of entertaining people, especially when all they really wanted were the same old tunes he had long ago become bored by". While he was in retirement, President Richard Nixon asked him to perform at a Young Voters Rally in anticipation of the upcoming campaign. \hl{Sinatra} obliged and chose to sing "My Kind of Town" for the rally held in Chicago on October 20, 1972. In 1973, \hl{Sinatra} came out of his short-lived retirement with a television special and album. The album, entitled \hl{"Ol' Blue Eyes} Is Back", arranged by Gordon Jenkins and Don Costa,\\
- - - - - - - - - - - - - - - - - - - - - - - - - - - - - - - - - - - - - - - - - - - - -  \\
\textbf{AMR graph information:} \# nodes: 52,  \# edges 85, \# SSSP: 19\\
\textbf{Score by palm~2: }20/100,  \textbf{Rank by palm~2: }27/100\\
\textbf{Rank by DPR:} 8/100
\end{tcolorbox}
\begin{tcolorbox}
\textbf{4th:} State Police would attend, searching for organized crime members in the audience. During a 1979 appearance in Providence, Mayor Buddy Cianci named Sinatra an honorary fire chief, complete with a helmet bearing the name "F. \hl{SINATRA}" with \hl{nickname "Ol' Blue Eyes"} beneath. David Bowie's concert on May 5, 1978 was one of three recorded for his live album "Stage". The Bee Gees performed two sold-out concerts here on August 28–29, 1979 as part of their Spirits Having Flown Tour. The Kinks recorded much of their live album and video, "One for the Road" at the Civic Center September 23, 1979.\\
- - - - - - - - - - - - - - - - - - - - - - - - - - - - - - - - - - - - - - - - - - - - -  \\
\textbf{AMR graph information:} \# nodes: 54,  \# edges 67, \# SSSP: 0\\
\textbf{Score by palm~2: }50/100,  \textbf{Rank by palm~2: }9/100\\
\textbf{Rank by DPR:} 6/100

\end{tcolorbox}

\begin{tcolorbox}
\textbf{5th:} illness). Pasetta was the producer of the Elvis Presley concert special, "Aloha from Hawaii Via Satellite" in January 1973. The show still holds the record for the most watched television special in history; viewing figures are between 1 and 1.5 billion live viewers worldwide. 1973 also saw Pasetta direct "Magnavox Presents \hl{Frank Sinatra}" (also known as \hl{"Ol' Blue Eyes Is Back"}), the television special that marked Frank Sinatra's comeback from retirement. Pasetta died in a 2015 single-car accident. The vehicle driven by Keith Stewart collided with Pasetta shortly after Stewart had allowed his passengers to disembark. Marty Pasetta Martin Allen\\
- - - - - - - - - - - - - - - - - - - - - - - - - - - - - - - - - - - - - - - - - - - - -  \\
\textbf{AMR graph information:} \# nodes: 39,  \# edges 59, \# SSSP: 39\\
\textbf{Score by palm~2: }50/100,  \textbf{Rank by palm~2: }9/100\\
\textbf{Rank by DPR:} 3/100

\end{tcolorbox}

\begin{tcolorbox}
\textbf{6th:} 
him feel wealthy and important, and that he was giving his very best to the audience. He was also obsessed with cleanliness—while with the Tommy Dorsey band he developed the nickname "Lady Macbeth", because of frequent showering and switching his outfits. His deep blue eyes earned him the popular \hl{nickname "Ol' Blue Eyes"}. For Santopietro, Sinatra was the personification of America in the 1950s: "cocky, eye on the main chance, optimistic, and full of the sense of possibility". Barbara \hl{Sinatra} wrote, "A big part of Frank's thrill was the sense of danger that he exuded, an underlying, ever-present tension only\\
- - - - - - - - - - - - - - - - - - - - - - - - - - - - - - - - - - - - - - - - - - - - -  \\
\textbf{AMR graph information:} \# nodes: 44,  \# edges 81, \# SSSP: 30\\
\textbf{Score by palm~2: }100/100,  \textbf{Rank by palm~2: }1/100\\
\textbf{Rank by DPR:} 2/100
\end{tcolorbox}

\begin{tcolorbox}
\textbf{7th:} 
where his suite and those of his entourage were on the 23rd floor. His tour, his first in Australia in 15 years and billed as \hl{"Ol' Blue Eyes} Is Back," was scheduled to include two shows in Melbourne, followed by three in Sydney. In his first show, according to news reports from 1974, \hl{Sinatra} referred on stage to the media as "parasites" and "bums" and to women specifically as "the broads of the press, the hookers of the press," then adding, "I might offer them a buck and a half, I'm not sure." The character of Rod Blue in the\\
- - - - - - - - - - - - - - - - - - - - - - - - - - - - - - - - - - - - - - - - - - - - -  \\
\textbf{AMR graph information:} \# nodes: 32,  \# edges 61, \# SSSP: 32\\
\textbf{Score by PaLM~2: }50/100,  \textbf{Rank by palm~2: }9/100\\
\textbf{Rank by DPR:} 11/100

\end{tcolorbox}

\begin{tcolorbox}
\textbf{8th:} RLPO, BBC Concert Orchestra (for "Friday Night Is Music Night"), Lahti Symphony Orchestra, Northern Sinfonia, the Melbourne Symphony Orchestra, the Adelaide Symphony Orchestra for the Adelaide Cabaret Festival and the RTÉ Concert Orchestra. His most popular show is the interactive "\hl{Sinatra} Jukebox" where, "instead of an hour of songs and anecdote, halfway through members of the audience were invited to fill in request forms". Reviewing the show, "Cabaret Scenes" said, "I can think of no other singer to better pay homage to \hl{Ol' Blue Eyes} on his 100th birthday." In 2014 he performed on BBC Radio 2 with the BBC\\
- - - - - - - - - - - - - - - - - - - - - - - - - - - - - - - - - - - - - - - - - - - - -  \\
\textbf{AMR graph information:} \# nodes: 51,  \# edges 57, \# SSSP: 20\\
\textbf{Score by palm~2: }20/100,  \textbf{Rank by palm~2: }27/100\\
\textbf{Rank by DPR:} 14/100

\end{tcolorbox}
\begin{tcolorbox}
\textbf{9th:} 
as his tribute to "The Great American Songbook". The album has Oleg’s vocals and arrangements by a big band leader Patrick Williams (a late period \hl{Frank Sinatra} recording associate) and sound engineering by Al Schmitt, whose 60-year career yielded 150 gold and platinum albums, 20 Grammy awards and who also recorded \hl{Ol' Blue Eyes}. "Bring Me Sunshine" was produced at the legendary Capitol Records studios in Hollywood, CA. Songwriter, Charles Strouse quoted: ""The Great American Songbook, to which I am proud to be a contributor, is one of our greatest cultural exports, Oleg is a living example of what an\\
- - - - - - - - - - - - - - - - - - - - - - - - - - - - - - - - - - - - - - - - - - - - -  \\
\textbf{AMR graph information:} \# nodes: 52,  \# edges 68, \# SSSP: 12\\
\textbf{Score by palm~2: }50/100,  \textbf{Rank by palm~2: }9/100\\
\textbf{Rank by DPR:} 27/100
\end{tcolorbox}

\begin{tcolorbox}
\textbf{10th:} 
first place wins The Founding Director is Ben Ferris (2004+). Sydney Film School runs two courses: The Diploma of Screen \& Media and The Advanced Diploma of Screen \& Media. Some of the accolades afforded to Sydney Film School graduates for their work include: Best Student Documentary Film at Antenna Film Festival: \hl{"Ol' Blue Eyes"}, Matt Cooney Finalist at Bondi Short Film Festival: "Letters Home", Neilesh Verma Industry Advisory Board (IAB) Pitch Competition winner: "Lotus Sonny", Gary Sofarelli Opening Night screening; Best Australian Animation \& Best Australian Composer at World of Women WOW Film Festival, 2012: "Camera Obscura", Marta Maia\\
- - - - - - - - - - - - - - - - - - - - - - - - - - - - - - - - - - - - - - - - - - - - -  \\
\textbf{AMR graph information:} \# nodes: 58,  \# edges 69, \# SSSP: 35\\
\textbf{Score by palm~2: }0/100,  \textbf{Rank by palm~2: }30/100\\
\textbf{Rank by DPR:} 39/100

\end{tcolorbox}
By analyzing the above result, we note that documents (such as 1st, 2nd, and 4th) containing exact words from the question (i.e., these words are ``Ol' Blue Eyes" and  ``nickname" in our example) are prioritized at the top by most rankers. However, if a document includes word variations or lacks sufficient keywords, it poses a challenge for the baseline reranker to identify its relevance, see the 9th and 10th documents. To address this issue, the AMR graph of documents is used in our method to comprehend more intricate semantics. The SSSPs from the `question' node in the AMR graph also play the crucial role in uncovering the underlying connections between the question and the words in the documents.

Another challenging scenario for the baseline reranker arises when several keywords or even gold answers are present in the documents but are weakly connected, making recognition difficult. For example, in the 7th, 8th, and 9th documents there are both ``Ol' Blue Eyes" and ``Sinatra" which are gold answers, yet these words are not directly linked as the sentence: ``Ol' Blue Eyes is the nickname of ``Sinatra". Instead, the connection between these two words is very loose. Luckily, the 7th, 8th, and 9th documents are connected to the 1st document in the document graph due to common nodes like 'Sinatra' and 'Ol Blue Eyes.' The 1st document stands out as more easily identifiable as a positive document, given its incorporation of all keywords from the questions. These words not only have a strong connection but also collectively contribute to a cohesive answer to the question. Leveraging this information and employing a message-passing mechanism, we can enable the 7th, 8th, and 9th document to adeptly discern potential keywords. Consequently, this approach enhances their ranking, based on the insights derived from the well-connected and information-rich 1st document.

\section{Examples of LLM-generate Relevant Score}
\begin{figure}[htp]
    \centering
    \includegraphics[width=\textwidth]{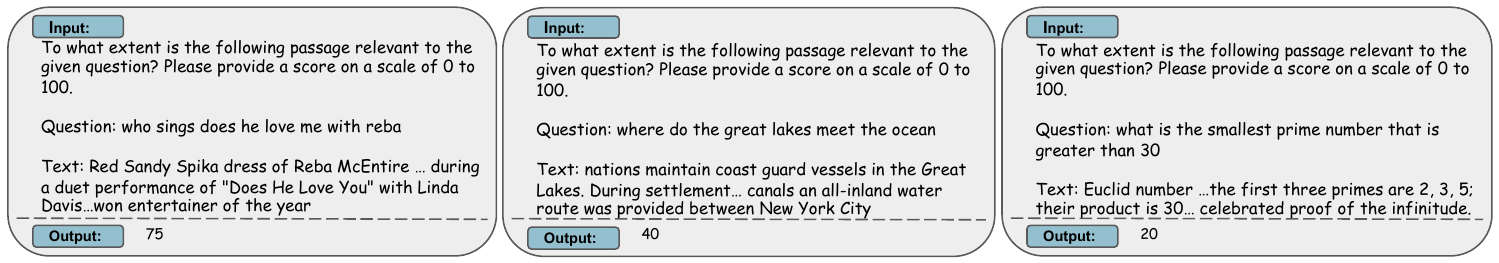}
    \caption{Examples of LLM-generate relevant score.}
    \label{fig:ulm_example}
\end{figure}

Some examples of LLM-generate relevant score are illustrated in Fig \ref{fig:ulm_example}.
 \end{document}